\documentclass[final,5p,times,twocolumn]{elsarticle}

\usepackage{amssymb}
\usepackage{amsmath}
\usepackage{caption3}

\usepackage{amssymb}
\usepackage{graphicx}
\usepackage{multirow}
\usepackage{longtable}
\usepackage{rotating}
\usepackage{color}
\usepackage{makecell}
\usepackage{hyperref}

\usepackage{bm}
\usepackage{lineno}
\usepackage{bm}
\usepackage{mathrsfs}
\usepackage{amsmath}
\usepackage{amssymb}
\usepackage{algorithm}
\usepackage{algorithmic}
\usepackage{array}
\usepackage{bm}
\usepackage{booktabs}
\usepackage{graphicx}
\usepackage{bigstrut}
\usepackage{caption3}
\usepackage{booktabs}
\usepackage{colortbl}
\usepackage{url}
\usepackage{multirow}
\usepackage{xcolor}
\usepackage{bbding}

\journal{Nuclear Physics B}

\begin{document}

\begin{frontmatter}



\title{GrFormer: A Novel Transformer on Grassmann Manifold for Infrared and Visible Image Fusion}


\author[1]{Huan~Kang}
\author[1]{Hui~Li}
\author[1]{Xiao-Jun Wu\corref{a}\corref{correspondingauthor}}
\author[1]{Tianyang~Xu}
\author[1]{Rui~Wang}
\author[1]{Chunyang~Cheng}
\author[2]{Josef~Kittler}

\address[1]{ School of Artificial Intelligence and Computer Science, Jiangnan University,  214122, Wuxi, China 
}

\address[2]{The Centre for Vision, Speech and Signal Processing, University of Surrey, GU2 7XH, Guildford,  UK }

\cortext[correspondingauthor]{Corresponding author email: wu\_xiaojun@jiangnan.edu.cn}

\begin{abstract}
In the field of image fusion, promising progress has been made by modeling data from different modalities as linear subspaces. 
However, in practice, the source images are often located in a non-Euclidean space, where the Euclidean methods usually cannot encapsulate the intrinsic topological structure. 
Typically, the inner product performed in the Euclidean space calculates the algebraic similarity rather than the semantic similarity, which results in undesired attention output and a decrease in fusion performance. While the balance of low-level details and high-level semantics should be considered in infrared and visible image fusion task. 
To address this issue, in this paper, we propose a novel attention mechanism based on Grassmann manifold for infrared and visible image fusion (\textbf{\textit{GrFormer}}). Specifically, our method constructs a low-rank subspace mapping through projection constraints on the Grassmann manifold, compressing attention features into subspaces of varying rank levels. This forces the features to decouple into high-frequency details (local low-rank) and low-frequency semantics (global low-rank), thereby achieving multi-scale semantic fusion.
Additionally, to effectively integrate the significant information, we develop a cross-modal fusion strategy (CMS) based on a covariance mask to maximise the complementary properties between different modalities and to suppress the features with high correlation, which are deemed redundant. 
The experimental results demonstrate that our network outperforms SOTA methods both qualitatively and quantitatively on multiple image fusion benchmarks. The codes are available at \href{https://github.com/Shaoyun2023}{https://github.com/Shaoyun2023}.
\end{abstract}




\end{frontmatter}



\section{Introduction}
\label{sec:intro}
Image fusion is a technique that integrates complementary information from multiple sensors or diverse imaging conditions to generate a unified, comprehensive representation of the scene. By leveraging the distinct yet complementary characteristics of different modalities~\cite{ma2019infrared,zhang2021image}—such as thermal radiation in infrared imaging and texture details in visible light, this technology produces fused images with enhanced informational content and improved visual interpretability, thereby facilitating more accurate scene understanding and analysis.
\begin{figure}
    \centering
    \includegraphics[width=1\linewidth]{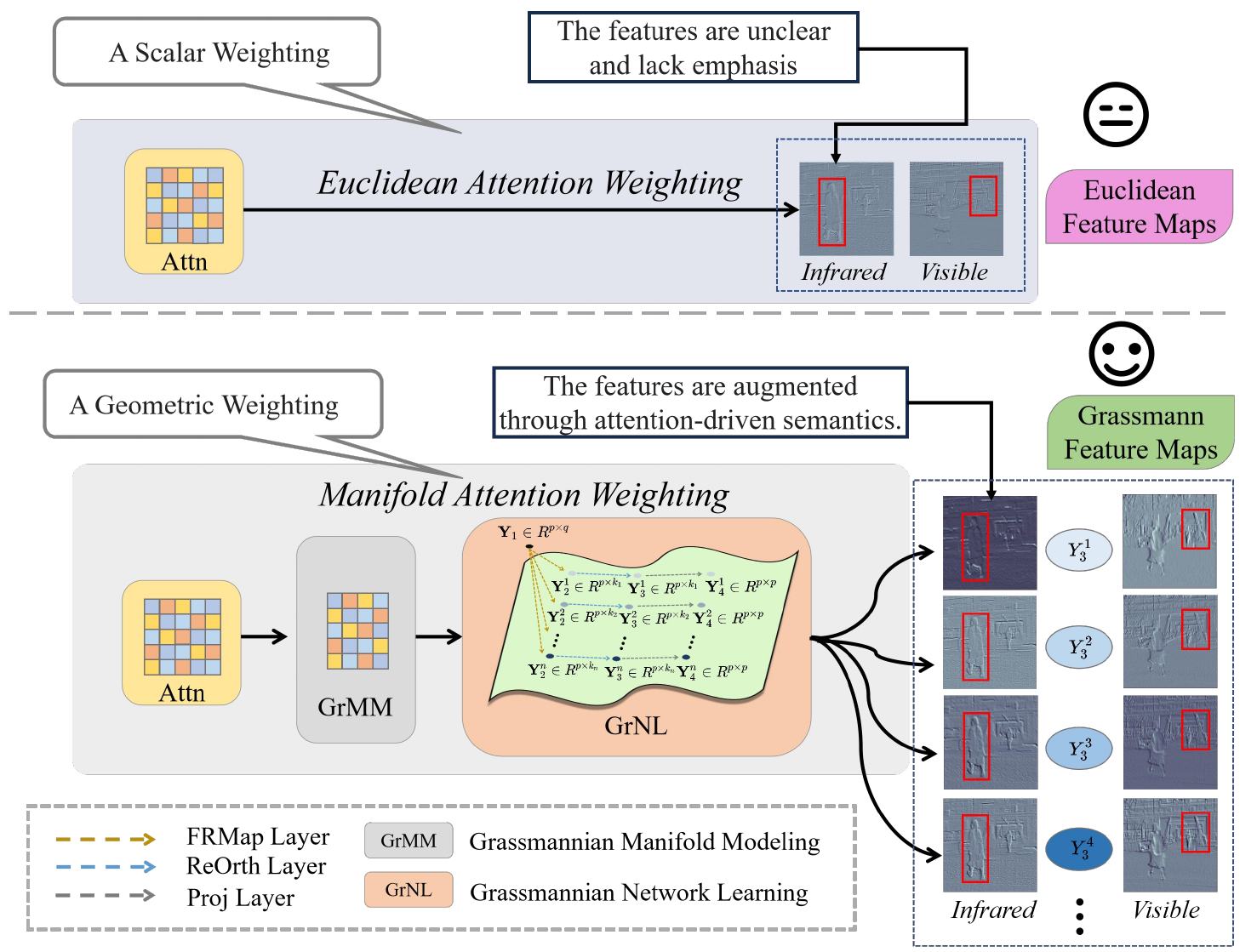}
    \caption{Previous Euclidean-based methods perform algebraic attention weighting, which can inadvertently weaken feature representations. In contrast, our approach emphasizes semantic similarity computation, adhering to the geometric structure of the Grassmann manifold for modeling. By decomposing high-frequency and low-frequency information across modalities, our method generates more reasonable and discriminative attention outputs.}
    \label{summary}
\end{figure}
For example, in the infrared and visible image fusion task~\cite{jiang2021review,xu2020u2fusion}, visible sensors excel in preserving texture details and vivid colors but may fail to capture the key information in low light conditions or when obstructions appear.
Conversely, although infrared images cannot preserve these fine-grained texture details, they can highlight targets under these adverse conditions.
Fusing these two modalities together, we can obtain high-quality images that preserve both rich texture details and salient thermal information, thus overcoming the imaging limitations of sensors.
This technique plays a significant role in various fields such as autonomous driving~\cite{prakash2021multi}, medical imaging~\cite{tang2022matr,zhou2023gan}, visual object tracking~\cite{tang2023exploring,zhu2022visual}, etc.

However, effectively integrating these complementary modalities requires sophisticated mechanisms to resolve inherent discrepancies in cross-modal representations, such as the attention mechanism~\cite{dosovitskiy2020vit,hu2018squeeze,vaswani2017attention,wang2018non,woo2018cbam,zhang2019self}. It originally emerged from cognitive science in the 1990s and quickly expanded into the field of computer vision, simultaneously driving the development of multimodal learning.
The spatial and channel attention mechanisms enable fusion models to dynamically allocate weights based on the content of the input images~\cite{li2020nestfuse,rao2023tgfuse,xiao2020global}, improving the extraction of salient image features.
However, these methods often prioritize intra-modal feature associations but overlook inter-modal relationships, which are essential for fusion tasks. We argue that the complementary information of different modalities should be emphasized more vigorously by enhancing the internal features with low correlation.
In recent research, some methods have recognised this issue and designed approaches based on cross-attention~\cite{jia2023multiscale,li2024crossfuse,kang2024spdfusion}, which delivers promising results.
While cross-attention improves interaction, existing methods still struggle to fully decouple modality-specific features and efficiently model high-dimensional geometric relationships.
To address this research gap, we propose a manifold learning framework. Unlike conventional approaches that rely solely on Euclidean metrics, our method embeds high-dimensional data into the Grassmann manifold, effectively preserving local Euclidean relationships while capturing global nonlinear correlations.
Specifically, the geometric structure of the Grassmann manifold inherently facilitates cross-modal feature decoupling through its orthonormal basis system. When processing infrared and visible images, this architecture automatically separates spectral and textural features into distinct yet geometrically coherent subspaces via orthogonal matrix mappings, thereby maintaining inter-modal information integrity, and this proves particularly crucial for infrared-visible image fusion tasks.
By leveraging the manifold's intrinsic properties, our framework provides a more natural representation for multimodal fusion, overcoming the limitations of purely Euclidean-based approaches.

Therefore, in this paper, we propose a novel transformer architecture for infrared and visible image fusion based on the Grassmann manifold (\textbf{\textit{GrFormer}}), which achieves semantic similarity computation within the attention mechanism to extract meaningful image outputs. As shown in Fig. \ref{summary}, we dynamically enhance semantic alignment and salient region fusion of cross-modal features through the Grassmann manifold-constrained attention mechanism, achieving complementary interaction between infrared and visible features in the low-rank manifold space. Compared to Euclidean attention, the features we obtain contain richer semantic information. At the same time, the Riemannian manifold-based attention network is extended to the spatial domain to better adapt to the intrinsic topological structure of the image, thereby achieving a better fusion of infrared and visible images.
Finally, a new cross-modal strategy is introduced for the cross-modal image fusion task.
The proposed embedding of this module into the classical Vision Transformer (ViT) structure~\cite{dosovitskiy2020vit} highlights the low-correlation (complementary) features of the two modalities and facilitates the analysis of inter-modal statistical relationships. The main contributions of this paper are summarized as follows:

\begin{itemize}
    \item We propose a novel model to embed Grassmann subspace representations in Euclidean attention networks, which first extends both spatial and channel attention mechanisms to manifold space for fusion. This approach effectively captures and fuses the inherent structural and semantic properties of multi-modal image data.
    \item Our framework constructs low-rank subspace mappings to disentangle high-frequency details and low-frequency semantics in images, enabling hierarchical cross-modal statistical relation learning through manifold-constrained optimization.
    \item We propose a novel fusion strategy that leverages learnable mask tensors to achieve foreground-background separation, effectively enhancing complementary information exchange across modalities.
    \item Experimental results on widely used benchmarks clearly demonstrate the superior performance of the proposed manifold-based approach, both qualitatively and quantitatively.
    
\end{itemize}

\section{Related work}
\label{sec:formatting}
In this section, we first review classic deep learning fusion frameworks, followed by an overview of feature decomposition-based fusion methods. We then present a detailed discussion on Grassmann manifold subspaces and their relevance to our work.

\begin{figure*}
    \centering
    \includegraphics[width=1\linewidth]{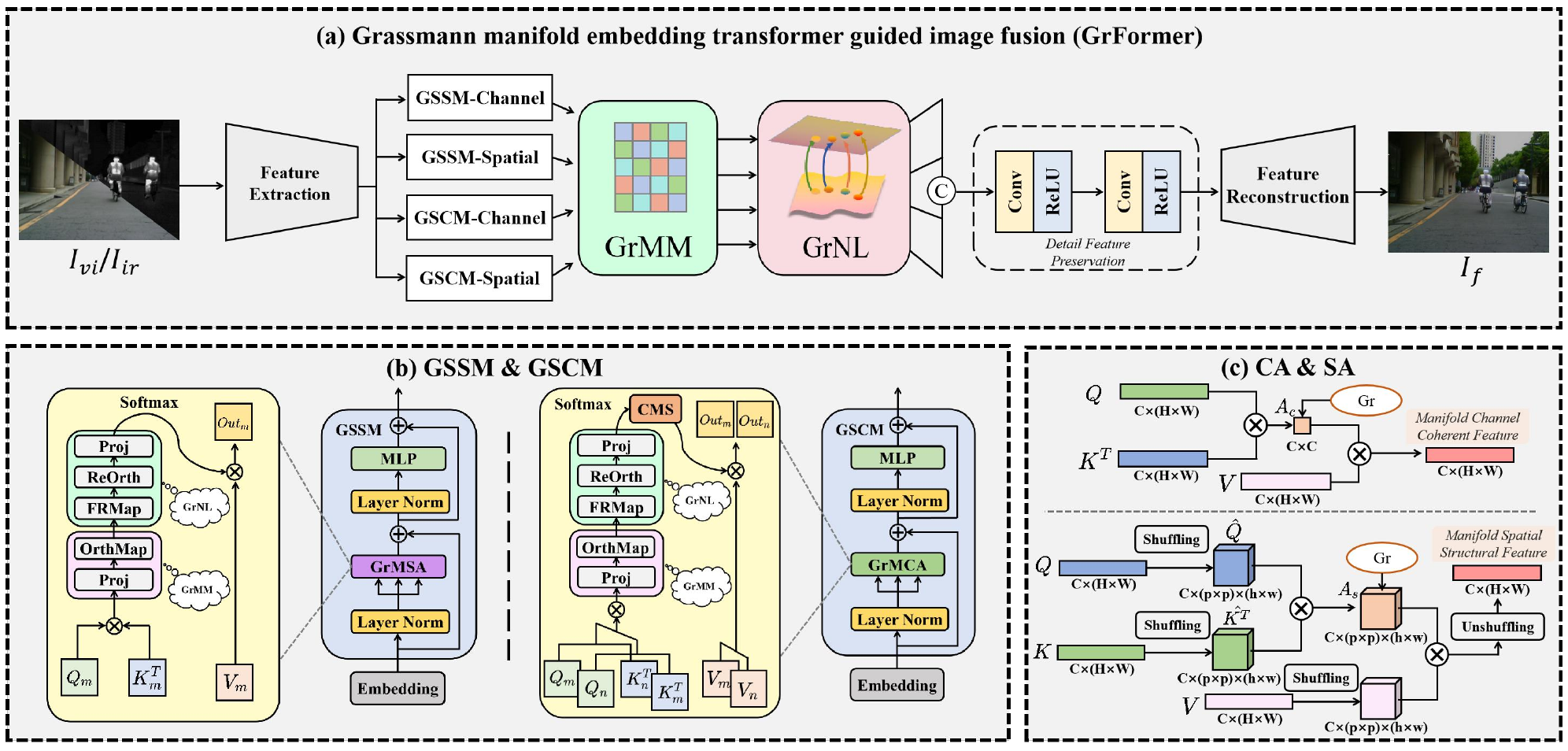}
    \caption{The workflow of our GrFormer. In the encoding stage, the input is first encoded by convolutional layers and divided into patches, followed by processing through the Grassmann-embedded self-attention module (GSSM) and cross-attention module (GSCM), which effectively capture both intra-modal and inter-modal discriminative semantic information. To achieve a more comprehensive information representation, we extend these two network architectures to both spatial and channel dimensions and integrate them through concatenation. In the decoding stage, a convolutional network-based decoder is employed to generate the fused image. In (c), ``Gr" denotes both Grassmann manifold modeling (GrMM) and Grassmann network learning (GrNL). Specifically, the ``GSCM" incorporates an additional cross-modal fusion strategy (CMS).}
    \label{workflow of GrFormer}
\end{figure*}

\subsection{Fusion methods based on deep learning}
Previous fusion networks have achieved impressive results by leveraging the powerful fitting capabilities of deep learning. These methods utilise Convolutional Neural Networks (CNNs) for efficient feature extraction and reconstruction ~\cite{li2018densefuse,li2020nestfuse,zhang2020ifcnn}. However, the quality of fusion results heavily depends on handcrafted fusion strategies. Consequently, some end-to-end deep networks~\cite{li2021rfn,xu2020u2fusion} are proposed to address this issue. Recent work further tackles misaligned inputs by integrating implicit registration with fusion in a unified framework ~\cite{li2024deep,li2025mulfs,tang2022superfusion}, eliminating reliance on pre-alignment. Meanwhile, generative paradigms including GANs ~\cite{ma2019fusiongan,ma2020ddcgan} and meta-learning-enhanced architectures ~\cite{zhao2023metafusion} have demonstrated advantages in texture generation and modality adaptation. These advances collectively underscore the evolving paradigms in feature representation and fusion strategy learning.

In addition, the introduction of attention mechanisms has significantly accelerated the advancement of image fusion. Some CNN-based fusion methods demonstrated attention's effectiveness through dual mechanisms: ~\cite{tang2023dual} combined channel and spatial attention for adaptive feature fusion, while ~\cite{li2020nestfuse} used nested connections with multi-scale attention to preserve critical information. These studies established an important foundation for applying attention mechanisms to fusion tasks.

Building upon these foundations, the field has witnessed a paradigm shift with the introduction of Transformer architectures. Recent Transformer-based methods ~\cite{ma2022swinfusion,zhao2023cddfuse,li2024crossfuse,qu2022transmef,rao2023tgfuse} have advanced fusion performance through self-attention mechanisms, effectively capturing global dependencies while preserving modality-specific features. These approaches excel in tasks that require precise spatial alignment across imaging modalities.  

Unfortunately, this global attention mechanism may overlook the low-rank structure of image regions, resulting in insufficient capture of local details, which affects the fusion performance.

\subsection{Feature decomposition-based fusion methods}
The field of infrared and visible image fusion has witnessed significant advances through diverse methodologies that decompose and integrate multimodal information in distinct yet complementary ways. Among these, STDFusionNet\cite{ma2021stdfusionnet} and SMR-Net\cite{SMRNet} explicitly decompose salient targets and texture details, using spatial masks to guide the fusion process, while SSDFusion\cite{ming2025ssdfusion} further decomposes images into scene-related and semantic-related components, enriching contextual information by injecting fusion semantics. While these methods have explored salient target preservation and scene-semantic decomposition respectively, they fundamentally operate within Euclidean space and rely on implicit feature separation. In contrast, our Grassmann manifold-based framework explicitly models cross-modal relationships through geometric priors, eliminating the need for heuristic masking or manual decomposition. Similarly, FAFusion\cite{xiao2024fafusion} decomposes images into frequency components to preserve structural and textural details but misses global nonlinear correlations in cross-modal data. To address this, our method leverages the Grassmann manifold's orthonormal basis to explicitly model these global nonlinear relationships, enabling more effective fusion.

These methods share a common underlying principle: the decomposition of multimodal data into interpretable components to facilitate effective fusion. This principle extends naturally to subspace-based methods, which operate on the assumption that data can be embedded into a low-dimensional subspace to capture its most significant features. Many fusion methods based on subspace representation have been proposed~\cite{li2020mdlatlrr,li2023lrrnet,zhang2021exploring}, leveraging the inherent structure of the data to identify and preserve critical information. Among the most commonly used are sparse and low-rank representation techniques~\cite{liu2012robust,wright2008robust}, which exploit both local and global structural properties of the data to extract features and conduct fusion more effectively. However, such linear subspace paradigms inherently disregard the nonlinear manifold geometry underlying multimodal imagery, where the geodesic consistency of intrinsic structures is critical for harmonizing low-level gradients with high-level semantics during fusion.

\subsection{Grassmann manifold subspace representation}
Over the past decade, tasks such as face recognition, skeleton-based action recognition, and medical image analysis have received a lot of attention.
Meanwhile, the learning method based on the Grassmann manifold representation has been widely applied in the classification task~\cite{huang2018building,sharma2020image,wang2023get,wang2020graph}.

GrNet~\cite{huang2018building} first generalises Euclidean neural networks to the Grassmann manifold, marking a novel exploration of deep network architecture. Following GrNet, GEMKML~\cite{wang2020graph} realises video frame sequence classification by constructing a lightweight cascaded feature extractor to hierarchically extract discriminative visual information.
SRCDPC~\cite{sharma2020image} extends the research to affine subspaces, designing a new kernel function to measure the similarity between affine subspaces and generating a low-dimensional representation (RVF) of affine spaces through the diagonalization of the kernel-gram matrix.
Additionally, in~\cite{wang2023get} the authors integrate Riemannian SGD into the deep learning framework, enabling the simultaneous optimisation of class subspaces on the Grassmann manifold with other model parameters, thereby enhancing classification accuracy.

While other Riemannian manifolds have been explored for representation learning, they present certain limitations. For example, SPD manifolds, which model symmetric positive definite matrices (e.g., covariance descriptors via \( \mathcal{P}_n = \{X \in \mathbb{R}^{n \times n} | X=X^T, X \succ 0\} \))—excel in capturing second-order statistics but struggle with high-dimensional image data due to computational complexity and sensitivity to noise~\cite{harandi2014manifold,harandi2017dimensionality}. Similarly, Stiefel manifolds, defined as \( \mathcal{V}_{n,m} = \{X \in \mathbb{R}^{n \times m} | X^TX = I_m\} \), preserve orthonormality but enforce overly rigid constraints that may discard discriminative multi-modal correlations~\cite{nguyen2022deep}.

In contrast, Grassmann manifolds naturally encode affine-invariant subspace relationships. For example, while infrared and visible images may exhibit linear distortions due to sensor differences, their essential features (such as edge structures and thermal radiation distributions) correspond to subspaces that remain equivalent on the manifold~\cite{li2016multi}. This representation flexibly captures the underlying geometric structure of the data, complementing the long-range feature learning strengths of Transformers and making it particularly well-suited for multimodal fusion tasks.  

\section{Proposed method}
In this section, we provide a detailed description of our method.
The overall framework of this approach is presented in Fig. \ref{workflow of GrFormer}.
\subsection{Preliminaries}
The Grassmann manifold \(\mathcal{G}(q,d)\) consists of all \(q\)-dimensional linear subspaces in \(\mathbb{R}^{d}\), forming a compact Riemannian manifold of dimensionality \(q\left ( d-q \right ) \).
Each subspace is spanned by an orthonormal basis matrix \(\textbf{Y}\) with the dimension of \(d~\times q\), satisfying \(\textbf{Y}^{T}\textbf{Y}\) = \(\textbf{I}_{q}\), with \(\textbf{I}_{q}\) being the identity matrix of size \(q\times q\).
The projection mapping~\cite{edelman1998geometry} \(\Phi(\textbf{Y})=\textbf{Y}\textbf{Y}^{T}\) not only represents the linear subspace but also approximates the true geodesic distance on the Grassmann manifold.

\subsection{Image fusion pipeline}
In this section, we provide a detailed explanation of the pipeline for the Grassmann manifold embedded fusion network.
\subsubsection{Image encoder}
 In the initial stage of the network, \(I_{ir}\) and \(I_{vi}\) represent the infrared and visible images, respectively. Two separate streams are employed to process them individually, using  identical convolutional encoding layers to extract deep features \(\left \{ \Phi_{I}^{D},\Phi_{V}^{D} \right \} \) from the corresponding source images. This process is represented by \(\mathcal{D} \left (\cdot \right )\):
 \begin{equation}
    \label{deep feature}
     \Phi_{I}^{D}=\mathcal{D}(I_{ir}),\Phi_{V}^{D}=\mathcal{D}(I_{vi}).
 \end{equation}
 

\subsubsection{Grassmann manifold modeling}
In the Grassmann manifold attention module, we integrate a projection operation into the ViT architecture~\cite{dosovitskiy2020vit} to construct an attention-based fusion network on the Grassmann manifold, effectively leveraging low-rank semantics of distinct subspaces. Meanwhile, several common manifold operations will be defined in Section \ref{grassmann layer}. Let \(\textbf{X}_{\mathrm{k}}\in\mathbb{R}^{(h\times w) \times d}\) represent the input features. Here, \( \mathrm{k} \) indexes different modalities, while \( h \), \( w \), and \( d \) represent height, width, and number of channels, respectively. By learning \(d\)-dimensional projection matrices \(\textbf{W}\in\mathbb{R}^{d \times d}\) , we obtain the query, key, and value matrices respectively:
\begin{equation}
    \textbf{Q}=\textbf{X}_{\mathrm{k}}\textbf{W}_{\textbf{Q}},\textbf{K}=\textbf{X}_{\mathrm{k}}\textbf{W}_{\textbf{K}},\textbf{V}=\textbf{X}_{\mathrm{k}}\textbf{W}_{\textbf{V}}.
\end{equation}

To satisfy the orthogonality assumption of the queries and keys on the Grassmann manifold, we perform a projection operation on the attention matrix, as shown in Fig. \ref{workflow of GrFormer} (b):

\begin{equation}
    \mathcal{A}_{r}=\mathrm{OrthMap} (  \mathrm{Proj} ( \textbf{Q}^{T}\textbf{K} )  ),   
\end{equation}
where Proj~\cite{edelman1998geometry} is a projection mapping, OrthMap~\cite{huang2018building} is an orthogonal mapping layer on a Grassmann manifold, \(\mathcal{A}\) denotes the attention matrix and \(r\) is the index of subspace projection.

\subsubsection{Grassmann network learning}
Simultaneously, we project the attention matrix into different manifold subspaces, and use the classical Grassmann network to update parameters of the attention matrix:
\begin{equation}
    \mathcal{A}_{r}'=\mathrm{Proj} \left ( \mathrm{ReOrth\left ( \mathrm{FRMap\left ( \mathcal{A}_{\textit{r}} \right ) }    \right ) }  \right ),
\end{equation}

\begin{equation}
    \mathrm{Attention}(\textbf{Q},\textbf{K},\textbf{V})=\textbf{V}(\mathrm{softmax}(\frac{\mathrm{\mathcal{A}}_{r}' }{\sqrt{d_{inp}}})),
\end{equation}
where FRMap~\cite{huang2018building} is a full-rank mapping layer that projects attention features into multi-scale subspaces, separating high-frequency details from low-frequency semantics while adaptively preserving structural relationships. Meanwhile, ReOrth~\cite{huang2018building} is the re-orthonormalization layer, which subsequently enforces orthogonality via QR decomposition and inverse correction. The Proj maps subspace representations back to the original space via matrix projection, reconstructing the original attention matrix dimensionality. These operations collectively ensure geometric stability on the Grassmann manifold. The \(d_{inp}\) represents the dimension of the input vector.
The forward process of the Attention Module operation is as follows:
\begin{equation}
\begin{aligned}
&\textbf{X}_{\mathrm{k}}=\textbf{X}_{\mathrm{k}}+\mathrm{Attention}(\mathrm{Norm}(\textbf{X}_{\mathrm{k}})),\\
&\textbf{X}_{\mathrm{k}}=\textbf{X}_{\mathrm{k}}+\mathrm{MLP}(\mathrm{Norm}(\textbf{X}_{\mathrm{k}})),\\
&s.t.\ \ \mathrm{k}\in\left \{ ir,vi \right \} ,
\end{aligned}
\end{equation}

where \(\mathrm{Norm}(\cdot)\) denotes the normalization operation, and \(\mathrm{MLP}(\cdot)\) is a multi-layer perception.

\subsubsection{Grassmann-based transformer}
To fuse multi-modal features, we construct an attention network based on Grassmann manifold subspace. As shown in Fig. \ref{workflow of GrFormer} (a), GSSM and GSCM project the features onto subspaces through the FRMap layer and integrate information using the attention matrices.
We denote \(GSSM^{C} \left (\cdot \right )\) and \(GSSM^{S} \left (\cdot \right )\) as the Grassmann-based Transformers in channel and spatial domains of the intra-modality, respectively. Similarly, \(GSCM^{C} \left (\cdot \right )\) and \(GSCM^{S} \left (\cdot \right )\) represent the Grassmann-based Transformers in channel and spatial domains of the inter-modality.
We exchange queries between these two modalities, as is done in the cross attention mechanism (CAM).
The specific cross-modal fusion strategy is detailed in Section \ref{sec:mask}.
By manifold learning through four different spaces, we obtain low-rank features with statistical correlations within and across modalities \(\left \{ \Phi_{I,V}^{SM},\Phi_{I,V}^{CM} \right \} \), as well as the concatenated features \(\Phi_{I,V}^{C}\), which are defined as below:
\begin{equation}
    \Phi_{I,V}^{SM}=\left \{ GSSM^{C} \left (\Phi_{I,V}^{D} \right ),GSSM^{S} \left (\Phi_{I,V}^{D} \right ) \right \},
\end{equation}
\begin{equation}
    \Phi_{I,V}^{CM}=\left \{ GSCM^{C} \left (\Phi_{I,V}^{D} \right ),GSCM^{S} \left (\Phi_{I,V}^{D} \right ) \right \},
\end{equation}
\begin{equation}
    \Phi_{I,V}^{C}=\left \{ \Phi_{I,V}^{SM}, \Phi_{I,V}^{CM} \right \},
\end{equation}
where \(\Phi_{I,V}^{D}\) represents the depth features obtained by concatenating \(\Phi_{I}^{D}\) and \(\Phi_{V}^{D}\) in Equation \ref{deep feature}. ``\(\left \{ \right \}\)" is the channel concatenation operation.

\subsubsection{Fusion decoder}
In the decoder \(\mathcal{DC}(\cdot)\), features derived from manifold learning along the channel dimension serve as input. The fused image \(I_f\) is generated through a series of convolutional layers that progressively reduce dimensionality, thereby enhancing edge and texture preservation. Here, ``Feature Reconstruction" refers to the convolutional-layer-based fusion process that refines and integrates multi-source features into the final output.
The decoding process can be defined as:
\begin{equation}
    I_{f}=\mathcal{DC} \left (\Phi_{I,V}^{C} \right ).
\end{equation}

\begin{figure}[t]
    \centering
    \includegraphics[width=1\linewidth]{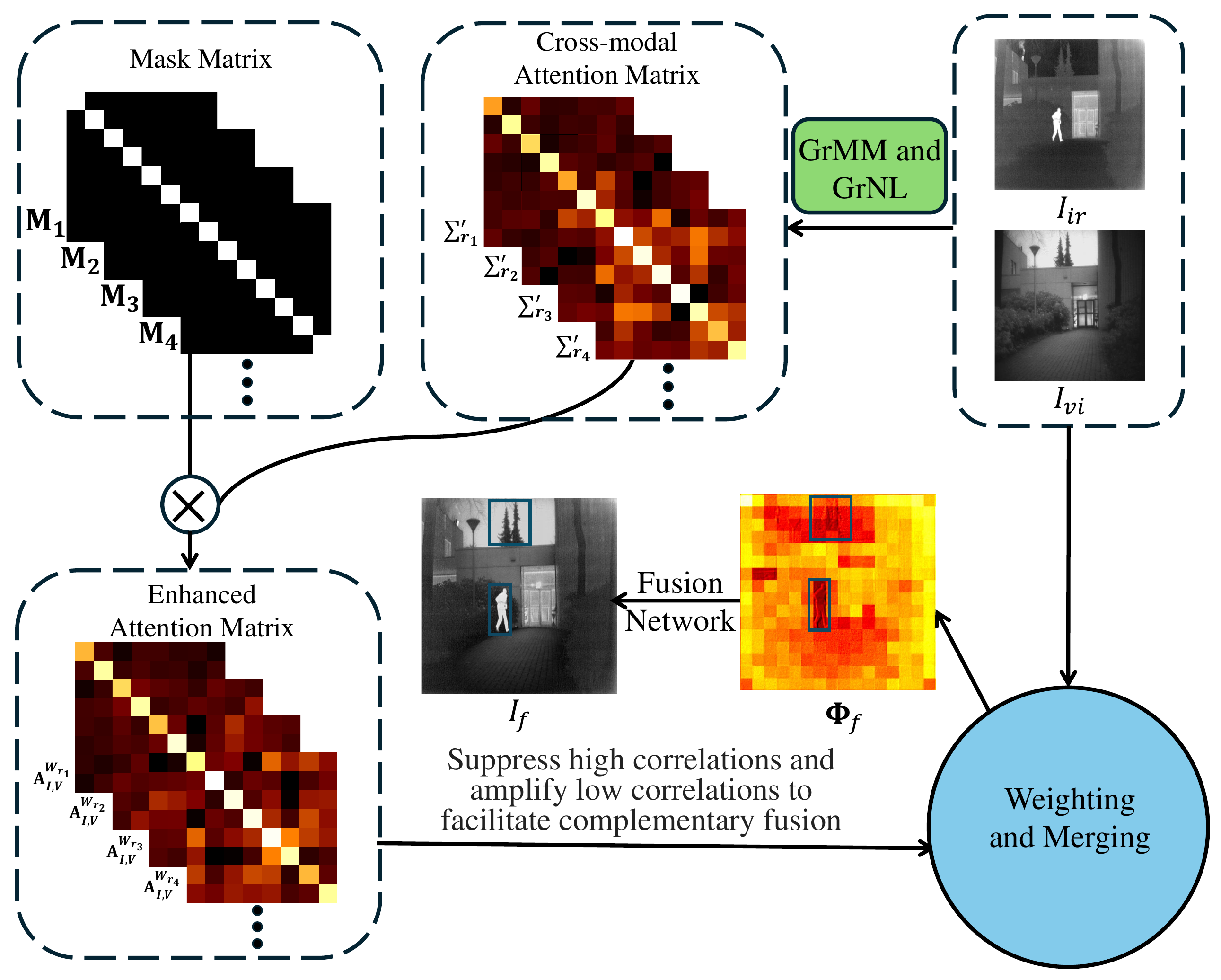}
    \caption{The framework of our cross-modal fusion strategy. It applies the mask matrix inside the covariance matrix to highlight the complementary information with low correlation and suppress the redundant information with strong correlation.}
    \label{strategy}
\end{figure}

\subsection{Grassmann manifold based attention}
\label{grassmann layer}
We replace the traditional scalar weighting with orthogonal transformations that conform to the Grassmann manifold. 

\subsubsection{OrthMap layer}
To ensure that the projected attention matrix satisfies the orthogonality constraint, we apply an OrthMap layer~\cite{huang2018building} to an attention matrix \(\textbf{Y}_{k-1}\) for the transformation:
\begin{equation}
     \textbf{Y}_{k} = f_{om}^{(k)}(\textbf{Y}_{k-1}) = \textbf{U}_{k-1,1:q},
\end{equation}
where \(k\) denotes the number of network layers, and \(\textbf{U}_{k-1,1:q}\) is obtained by performing eigenvalue (EIG) decomposition  \cite{ionescu2015training} on \(\textbf{Y}_{k-1}\) and extracting the first \(q\) largest eigenvectors.

\subsubsection{FRMap layer}
In the FRMap layer~\cite{huang2018building}, we aim to transform \(\textbf{Y}_{k}\) into a representation \(\textbf{Y}_{k+1}\) in a new space through a linear mapping. It is formulated as:
\begin{equation}
     \textbf{Y}_{k+1} = f_{fr}^{(k+1)}(\textbf{Y}_{k};\textbf{W}_{k+1}) = \textbf{W}_{k+1} \textbf{Y}_{k},
\end{equation}
where \(\textbf{W}_{k+1}\) is a transformation matrix that maps \(\textbf{Y}_{k}\) from the \( \mathbb{R}^{d_{k} \times q} \) space to the \( \mathbb{R}^{d_{k+1} \times q} \) space. Since \(\textbf{W}_{k+1}\) is row full-rank, this means it can preserve the structure of the subspace but may not remain the orthogonality.

\subsubsection{ReOrth layer}
Since \(\textbf{Y}_{k+1}\) may no longer be an orthogonal matrix, it is necessary to re-orthogonalize it using QR decomposition, and it is similar to the ReOrth layer~\cite{huang2018building}, \textit{i.e.}
\begin{equation}
      \textbf{Y}_{k+1} = \textbf{Q}_{k+1} \textbf{R}_{k+1},
\end{equation}
where \(\textbf{Q}_{k+1}\) is an orthogonal matrix and \(\textbf{R}_{k+1}\) is an upper triangular matrix. We then re-orthogonalize \(\textbf{Y}_{k+2}\) by the following manner:

\begin{equation}
      \textbf{Y}_{k+2} = f_{ro}^{(k+2)}(\textbf{Y}_{k+1}) = \textbf{Y}_{k+1} \textbf{R}_{k+1}^{-1}.
\end{equation}
In this way, \(\textbf{Y}_{k+2}\) is transformed back into an orthogonal matrix, preserving the orthogonality of the subspace.

\subsubsection{Projection layer}
To project \(\textbf{Y}_{k+2}\) into a lower-dimensional space and preserve its geometric structure, we construct a manifold layer based on projection operations:


\begin{equation}
      \textbf{Y}_{k+3} = f_{pm}^{(k+3)}(\textbf{Y}_{k+2}) = \textbf{Y}_{k+2} \textbf{Y}_{k+2}^T.
\end{equation}

The projection layer~\cite{huang2018building} uses linear transformations to expand the dimension of the orthogonal matrix, thereby reconstructing the attention weights in a manner that conforms the intrinsic relationships captured in the low-dimensional space.

\subsubsection{Spatial attention}
As illustrated in Fig. \ref{workflow of GrFormer} (c), to extend the manifold attention to the spatial dimension, we first reorganize input features through shuffling, which redistributes spatial elements into block-wise groupings. This step enables localized Grassmann low-rank projections and attention weighting to capture relationships between adjacent patches. After processing, an unshuffling operation restores the original spatial arrangement, ensuring global coherence while retaining attention-enhanced representations. It is worth noting that the QR decomposition significantly increases the computational complexity when dealing with multiple subspaces, making it necessary to seek an optimal trade-off between the algorithm efficiency and numerical robustness.
Thus, we select the most representative low-rank layers, through ablation experiments, as the feature representation of the manifold space attention.



\subsection{Cross-modal fusion strategy}\label{sec:mask}
The natural covariance matrix obtained by projecting onto the manifold through the Proj layer serves as our benchmark, which reflects the statistical correlation between patches of different modalities.
However, in image fusion tasks, regions with smaller correlations usually require more attention.
Thus, in our method, by adjusting the weights of the covariance matrix, we guide the network to focus on those complementary informations.
Fig. \ref{strategy} illustrates our strategy framework.

We treat the cross-modal attention matrix constructed from the images \(I_{ir}\) and \(I_{vi}\) as a metric tensor:

\begin{equation}
\textbf{M} = \left[ \begin{matrix}
1 & -1 & \cdots & -1 \\
-1 & 1 & \cdots & -1 \\
\vdots & \vdots & \ddots & \vdots \\
-1 & -1 & \cdots & 1
\end{matrix} \right]
.
\end{equation}

After the masking operation, we obtain the modality information-enhanced attention matrix \(\Sigma'_{r}\), where \(r\) represents the dimensionality of different subspaces:

\begin{equation}
\Sigma'_{r} = \textbf{M} \odot \Sigma_{r},
\end{equation}

where \textbf{M} represents the mask operation, and \(\Sigma_{r}\) denotes the original attention matrix.

Then, reshaped attention feature maps of different dimensions \(\textbf{A}_{I,V}^{W_{r}}\) are obtained by performing feature matrix-multiply between \(\Sigma '_{r}\) and \(\textbf{V}_{r}\). 

Finally, these feature maps are averaged and concatenated to obtain the fused feature map \(\Phi_{f}\):
\begin{equation}
    \Phi_{f}=\left \{ \frac{1}{r}\sum_{i=1}^{r}\textbf{A}_{I}^{W_{r}},\frac{1}{r}\sum_{i=1}^{r}\textbf{A}_{V}^{W_{r}} \right \}.
\end{equation}

This operation increases the ``distance" between different modal features, geometrically manifested as forcing data points to expand in directions with large inter-modal differences while maintaining the original structure within each modality.

\begin{figure*}[ht!]
\centering
\includegraphics[width=0.9\linewidth]{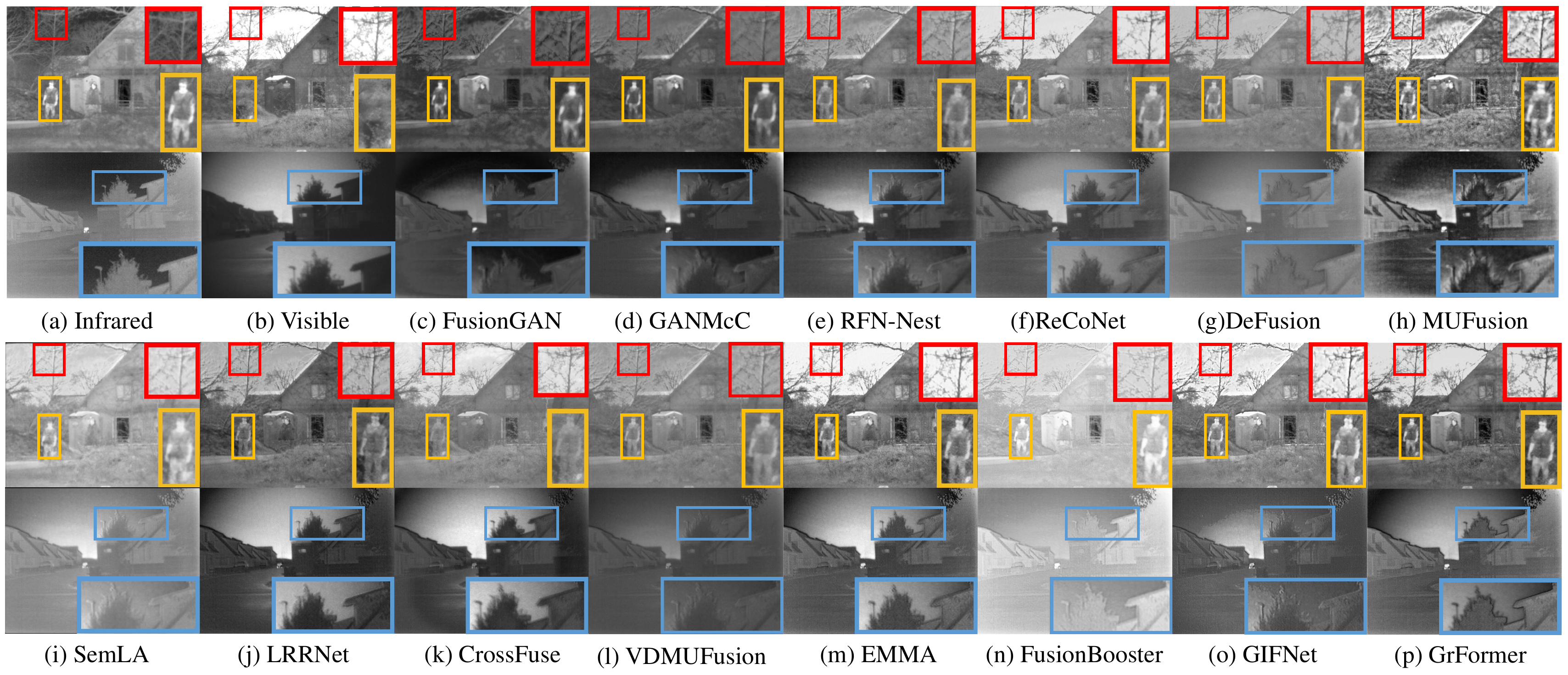}
\caption{Infrared and visible image fusion experiment on TNO dataset. The intricate semantic features of highly correlated regions are well-preserved, as exemplified by the distinct outlines of eaves and shrubs in the second and fourth rows. Simultaneously, complementary information from low-correlation regions is sufficiently emphasized, such as the contours of figures, the colors of clothing in the first and third rows, and the precise separation of tree branches from the sky background.} 
\label{TNO}
\end{figure*}

\begin{figure*}[ht!]
    \centering

        \centering
        \includegraphics[width=0.9\linewidth]{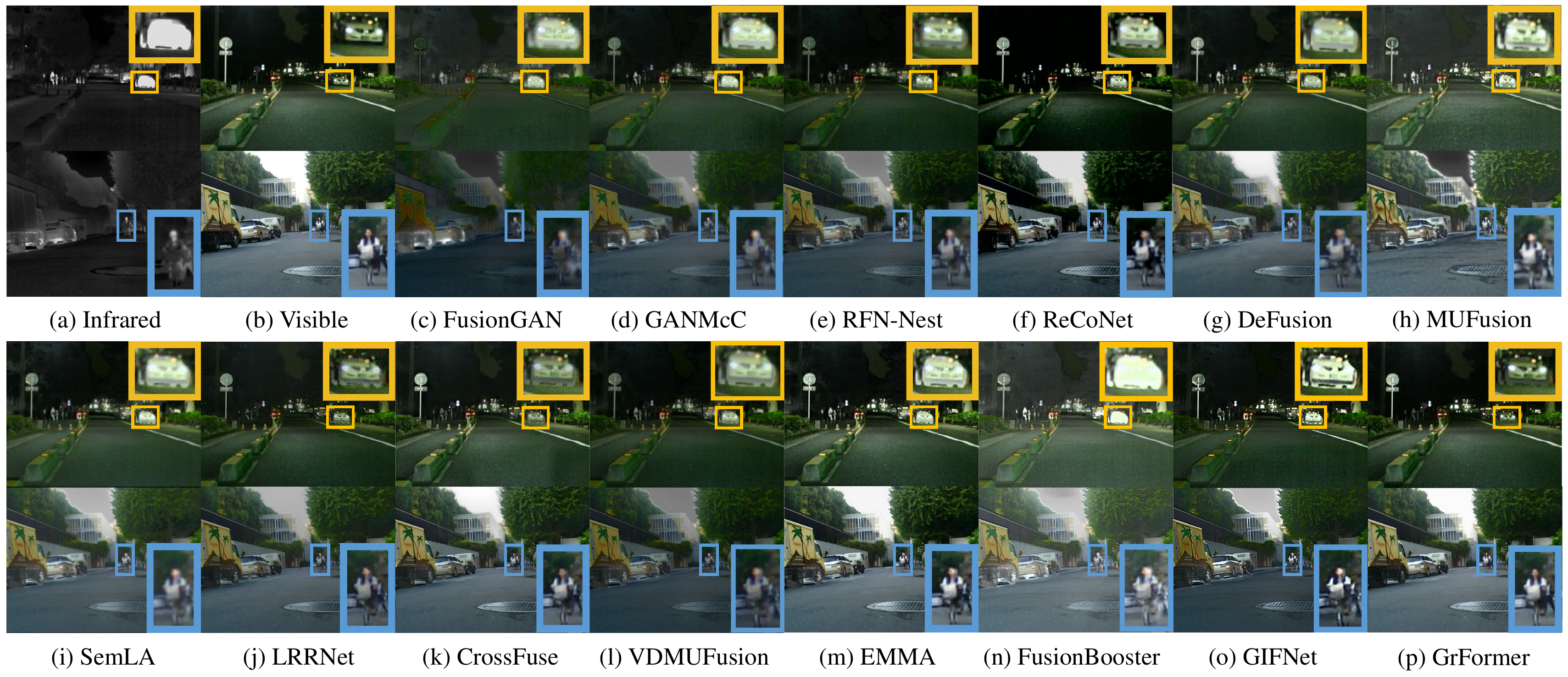}
        
        \label{fig:visible}

    \caption{Infrared and visible image fusion experiment on MSRS dataset. Our method effectively extracts the most valuable information from RGB images, as demonstrated in the first and third rows, where the details of the cars are more complete compared to other approaches. Simultaneously, in the second and fourth rows, the thermal infrared targets are prominently highlighted while effectively avoiding artifacts.}
    \label{MSRS}
\end{figure*}

\begin{table*}[ht!]
\centering 
\small
\captionsetup{justification=justified,singlelinecheck=false}
\caption{Quantitative Experiments on the TNO and MSRS Dataset. We represent the top three best-performing metrics using \textcolor{red}{\textbf{RED}}, \textcolor{brown}{\textbf{BROWN}}, and \textcolor{blue}{\textbf{BLUE}} fonts, respectively.}
\label{Quantitave TNO and MSRS}
\centering



\begin{tabular}{p{2.4cm}|p{0.6cm}p{0.7cm}p{0.7cm}p{0.6cm}p{0.8cm}p{0.9cm}|p{0.6cm}p{0.7cm}p{0.7cm}p{0.6cm}p{0.8cm}p{0.9cm}}
\toprule
& \multicolumn{6}{c|}{\textbf{TNO Dataset~\cite{TNO}}} & \multicolumn{6}{c}{\textbf{MSRS Dataset~\cite{tang2022piafusion}}}
\\
 & MI  & SF & VIF & AG & \(Q^{AB/F}\) & SSIM & MI  & SF & VIF & AG & \(Q^{AB/F}\) & SSIM
\end{tabular}

\begin{tabular}{p{2.4cm}|p{0.6cm}p{0.7cm}p{0.7cm}p{0.6cm}p{0.8cm}p{0.9cm}|p{0.6cm}p{0.7cm}p{0.7cm}p{0.6cm}p{0.8cm}p{0.9cm}} 


\midrule
 FusionGAN~\cite{ma2019fusiongan} & 2.330 & 6.306 & 0.413 & 2.422 & 0.230 & 0.332 & 1.893 & 4.354 & 0.442 & 1.452 & 0.140 & 0.242  \\
 GANMcC~\cite{ma2020ganmcc} & 2.276  & 6.189 & 0.520 & 2.553 & 0.277 & 0.424 & 2.565 & 5.664 & 0.635 & 2.006 & 0.302 & 0.393 \\
  RFN-Nest~\cite{li2021rfn} & 2.126 & 5.931 & 0.550 & 2.691 & 0.333 & 0.400 & 2.460 & 6.163 & 0.656 & 2.115 & 0.390 & 0.379\\
    ReCoNet~\cite{huang2022reconet} & 2.384 & 7.434 & 0.535 & 3.240 & 0.360 & 0.414 & 2.289 & 9.975 & 0.490 & 3.000 & 0.404 & 0.199 \\
 DeFusion~\cite{liang2022fusion} & 2.627 & 6.461 & 0.559 & 2.624 & 0.365 & \textcolor{brown}{\textbf{0.462}} & 2.990 & 8.146 & 0.730 & 2.654 & 0.507 & \textcolor{red}{\textbf{0.453}} \\

 MUFusion~\cite{cheng2023mufusion} & 1.965 & 10.374 & 0.539 & \textcolor{brown}{\textbf{4.923}} & 0.358 & 0.390 & 1.688 & 8.768 & 0.599 & 3.118 & 0.415 & 0.350 \\
 SemLA~\cite{xie2023semantics}  & 2.051 & 8.528 & 0.470 & 2.892 & 0.265 & 0.343 & 2.461 & 6.351 & 0.617 & 2.257 &  0.290 & 0.367 \\
 LRRNet~\cite{li2023lrrnet} & 2.518 & 9.608 & 0.548 & 3.790 & 0.352 & 0.406 & 2.928 & 8.473 & 0.541 & 2.651 & 0.454 & 0.211  \\
 CrossFuse~\cite{li2024crossfuse} & \textcolor{brown}{\textbf{2.952}} & 9.951 & \textcolor{brown}{\textbf{0.716}} & 3.733 & \textcolor{brown}{\textbf{0.425}} & \textcolor{blue}{\textbf{0.451}} & \textcolor{blue}{\textbf{3.132}} & 9.661 & \textcolor{blue}{\textbf{0.839}} & 3.019 & \textcolor{blue}{\textbf{0.561}} & \textcolor{brown}{\textbf{0.436}}\\
 VDMUFusion~\cite{shi2024vdmufusion} & 2.076 & 4.832 & 0.502 & 2.048 & 0.249 & 0.415 & 2.331 & 6.490 & 0.633 & 2.242 & 0.344 & 0.398 \\
 EMMA~\cite{zhao2024equivariant} & \textcolor{blue}{\textbf{2.717}} & \textcolor{brown}{\textbf{11.376}} & \textcolor{blue}{\textbf{0.600}} & \textcolor{blue}{\textbf{4.749}} & \textcolor{blue}{\textbf{0.405}} & 0.410 & \textcolor{brown}{\textbf{3.710}} & \textcolor{brown}{\textbf{11.565}} & \textcolor{brown}{\textbf{0.871}} & \textcolor{red}{\textbf{3.834}} & \textcolor{brown}{\textbf{0.573}} & \textcolor{brown}{\textbf{0.436}} \\
 FusionBooster~\cite{cheng2024fusionbooster} & 2.455 & 10.898 & 0.480 & 3.455 & 0.319 & 0.394 & 2.075 & 8.801 & 0.639 & 3.051 & 0.422 & 0.340 \\

 GIFNet~\cite{cheng2025cvpr_gifnet} & 1.990 & \textcolor{red}{\textbf{13.742}} & 0.501 & \textcolor{red}{\textbf{5.132}} & 0.347 & 0.433 & 1.971 & \textcolor{red}{\textbf{12.705}} & 0.582 & \textcolor{brown}{\textbf{3.367}} & 0.416 & \textcolor{blue}{\textbf{0.420}} \\
 \midrule
 GrFormer(Ours) & \textcolor{red}{\textbf{4.023}} & \textcolor{blue}{\textbf{11.100}} & \textcolor{red}{\textbf{0.873}} & 4.261 & \textcolor{red}{\textbf{0.538}} & \textcolor{red}{\textbf{0.481}} & \textcolor{red}{\textbf{3.952}} & \textcolor{blue}{\textbf{10.530}} & \textcolor{red}{\textbf{0.880}} & \textcolor{blue}{\textbf{3.207}} & \textcolor{red}{\textbf{0.574}} & \textcolor{brown}{\textbf{0.436}} \\ 
\bottomrule
\end{tabular}
\end{table*}

\subsection{Loss function}
The quality of the fused image is critically influenced by the design of the loss function. To facilitate the attention network in extracting rich, statistically relevant information from the source image across diverse intrinsic subspaces, we propose a detail-semantic complementary loss function. This loss function guides the network to effectively reconstruct the input modalities by balancing fine-grained details and high-level semantic features. 
The total loss function is defined as:
\begin{equation}
    L_{total}=L_{int}+\alpha L_{grad}+\beta L_{cov}+\gamma L_{ssim},
    \label{Ltotal}
\end{equation}
where \(L_{int}\) computes the \(l_{1}\) distance between the fused image and the element-wise maximum of the input images.
It is guided by reconstructing the source images at the pixel level to highlight the important regions.
Its definition is as follows:
\begin{equation}
    L_{int}=\frac{1}{HW}\parallel I_{f}-max(I_{ir},I_{vis}) \parallel _{1},
    \label{Lint}
\end{equation}
where \(H\) and \(W\) represent the height and width of an image, respectively. The \(max\left ( \cdot\right )\) function takes the maximum value of the corresponding elements in the input matrix, and \(\parallel \cdot \parallel _{1}\) is \(l_{1}-norm\).

To achieve a more precise texture representation in the subspace, we introduce gradient-based constraints between the source images and the fusion result, \textit{i.e.}, a set of regularization terms that minimize the discrepancies in gradient magnitudes and orientations:
\begin{equation}
    L_{grad}=\frac{1}{HW}\parallel \left | \nabla I_{f} \right | -max(\left | \nabla I_{ir} \right | , \left | \nabla I_{vis} \right | ) \parallel _{1},
    \label{Lgrad}
\end{equation}
where \(\nabla\) and \(|\cdot|\) represent the Sobel operator.

At the feature level, in order to maximize the retention of deep semantics in the feature subspace, we use the VGG-16 trained on ImageNet for feature extraction and select the deep convolutional blocks to design the loss function. The definition of \(L_{cov}\) is as follows:
\begin{equation}
    L_{cov}=\sum_{k=3}^{w} || Cov(\Phi (I_{f})^{k})-Cov(\Phi (I_{ir})^{k})||_{1},
    \label{Lcov}
\end{equation}
where \(Cov \left (\cdot \right )\) denotes the covariance matrix of the feature map and \(\Phi \left (\cdot \right )\) is the feature extracted from deep network. The \(w\) is set to 4.

Finally, we compute the structural similarity loss between the fused image and the source image to enforce structural consistency, defined as follows:
\begin{equation}
    L_{ssim}=(1-SSIM(I_{f},I_{vis}))+\delta (1-SSIM(I_{f},I_{ir})),
\end{equation}
where \(SSIM\) is the structural similarity index~\cite{wang2004image}, \(\delta\) is the balance term of loss.




\section{Experiments}
In this section, we introduce the implementation and configuration details, and validate the rationality of the proposed method and the effectiveness of the modules with experiments.
\subsection{Setup}
We first introduce the key components of the methodology, including the datasets used, parameter configurations, pipeline design, evaluation methods with quality metrics, and network optimization strategies.
\subsubsection{Datasets}
In our work, we selected 1083 pairs of corresponding infrared and visible images from the MSRS dataset as training data. During the testing phase, we use 40 pairs of images from TNO~\cite{TNO} and 361 pairs of images from MSRS~\cite{tang2022piafusion} as the test sets, respectively. The dimensions of the test images are typically not fixed.

\subsubsection{Parameter setting}
We implemented the algorithm using PyTorch. In the training phase, an end-to-end strategy was employed to train the model on an NVIDIA TITAN RTX GPU, and the size of the training images is standardized to $256\times 256$ pixels to ensure dimensional consistency across the network architecture. Within the manifold module, the Adam optimizer is used to update the weights of the Grassmann layers, with a learning rate set to \(10^{-4}\). Additionally, the parameters \(\alpha\), \(\beta\), \(\gamma\), and \(\delta\) in the loss function are empirically set to 1, 2, 10, and 1, respectively.

\subsubsection{Pipeline design}
The network employs a streamlined fusion architecture, first projecting inputs into higher dimensions via convolutional layers, then flattening the feature map into patches. Four parallel Grassmann manifold-based Transformer modules are used to process distinct attention types: (1) single-modal channel, (2) single-modal spatial, (3) cross-modal channel, and (4) cross-modal spatial attention. The feature dimension of the attention network is set to 64, with each attention head having a dimension of 8. The Channel Transformer assigns subspace coefficients to 2, 3, 4, 5 and aggregates features via summation, while the Spatial Transformer uses a fixed coefficient 100 for efficiency. Cross-modal interactions are explicitly modeled through dual-path attention between infrared/visible streams. During the decoding phase, all features are concatenated and progressively compressed via conv blocks (256→192→128→64→1 channels) to produce the fused output. The design unifies intra-modal relationships (channel/spatial), cross-modal interactions, and Grassmann manifold projections within a consistent framework. Notably, all convolutional blocks are configured with a kernel size of 3 and a stride of 1 to ensure consistency across the architecture.

\subsubsection{The methods and the quality metrics used}
The method presented in this article was compared and evaluated with thirteen different image fusion network approaches, including some classic and latest methods. These are:  FusionGAN~\cite{ma2019fusiongan}, GANMcC~\cite{ma2020ganmcc}, RFN-Nest~\cite{li2021rfn}, ReCoNet~\cite{huang2022reconet}, DeFusion~\cite{liang2022fusion}, MUFusion~\cite{cheng2023mufusion}, SemLA~\cite{xie2023semantics}, LRRNet~\cite{li2023lrrnet}, CrossFuse~\cite{li2024crossfuse}, VDMUFusion~\cite{shi2024vdmufusion}, EMMA~\cite{zhao2024equivariant}, FusionBooster~\cite{cheng2024fusionbooster} and GIFNet~\cite{cheng2025cvpr_gifnet}.
Regarding quality metrics, six indices were chosen for performance evaluation, which include: Mutual Information (MI), Spatial Frequency (SF), Visual Information Fidelity (VIF), Average Gradient (AG), \(Q^{AB/F}\) and structural similarity index measure (SSIM).
The descriptions of these metrics can be found in~\cite{ma2019infrared}.

\subsubsection{Network optimization}
As Grassmann optimization relies on EIG decomposition, we leverage the theoretical results of ~\cite{ionescu2015training} for gradient calculation. Consider the eigenvalue decomposition of a real symmetric matrix \( Y_{k-1} \in \mathbb{R}^{D \times D} \), where \( k \) denotes the layer number in the manifold network:
\begin{equation}
    Y_{k-1} = U \Sigma U^T ,
\end{equation}

where \( U \) is an orthogonal matrix (\( U^T U = I \)) and \( \Sigma \) is a diagonal matrix containing the eigenvalues. The gradient of the loss function \( L^{(k)} \) with respect to \( Y_{k-1} \) is derived as follows.

Under an infinitesimal perturbation \( dY_{k-1} \), the first-order variations of \( U \) and \( \Sigma \) are given by:

\begin{equation}
    d\Sigma = (U^T dY_{k-1} U)_{\text{diag}},
\end{equation}

\begin{equation}
    dU = U \left( \tilde{K} \circ (U^T dY_{k-1} U) \right),
\end{equation}

where \( \tilde{K} \) is the kernel matrix defined as:

\begin{equation}
    \tilde{K}_{ij} = \begin{cases} 
\frac{1}{\sigma_i - \sigma_j}, & i \neq j, \\
0, & i = j.
\end{cases}
\end{equation}

Here, \( \sigma_i \) denotes the \( i \)-th diagonal element of \( \Sigma \), and the gradient of \( L^{(k)} \) with respect to \( Y_{k-1} \) is obtained by applying the chain rule:

\begin{equation}
    \frac{\partial L^{(k)}}{\partial Y_{k-1}} : dY_{k-1} = \frac{\partial L^{(k)}}{\partial U} : dU + \frac{\partial L^{(k)}}{\partial \Sigma} : d\Sigma.
\end{equation}

Substituting the expressions for \( dU \) and \( d\Sigma \):

\begin{equation}
    \frac{\partial L^{(k)}}{\partial U} : dU = \left( \tilde{K} \circ \left( U^T \frac{\partial L^{(k)}}{\partial U} \right) \right) : (U^T dY_{k-1} U),
\end{equation}

\begin{equation}
    \frac{\partial L^{(k)}}{\partial \Sigma} : d\Sigma = \left( \frac{\partial L^{(k)}}{\partial \Sigma} \right)_{\text{diag}} : (U^T dY_{k-1} U).
\end{equation}

Combining these terms and projecting back to the \( Y_{k-1} \)-space using the orthogonal transformation property \( (U^T dY_{k-1} U) = U^T (\cdot) U \), we obtain:

\begin{equation}
\frac{\partial L^{(k)}}{\partial Y_{k-1}} = U \left[ \left( \tilde{K} \circ \left( U^T \frac{\partial L^{(k)}}{\partial U} \right) \right) + \left( \frac{\partial L^{(k)}}{\partial \Sigma} \right)_{\text{diag}} \right] U^T.
\end{equation}

\subsection{Comparison with SOTA methods}
In this section, we conducted both qualitative and quantitative experiments on the proposed GrFormer with two classic infrared-visible datasets, TNO and MSRS, to verify the performance of our method.



\subsubsection{Qualitative comparison}

In Fig. \ref{TNO} and Fig. \ref{MSRS}, we present the visualization results of four image pairs from two datasets. The comparative methods can be categorized into four groups. The first group consists of generative model-based approaches, including FusionGAN, GANMcC, and VDMUFusion. These methods tend to suppress outliers, causing high-brightness regions to be smoothed or compressed, resulting in darker fused images with reduced contrast, as seen in the sky in Fig. \ref{TNO}. The second group includes decomposition-based methods such as DeFusion, LRRNet, and FusionBooster. Due to lightweight autoencoders or low-rank constraints compressing feature dimensions, high-frequency details are lost, exemplified by the texture of trees in Fig. \ref{TNO}. The third group comprises training-based methods, including RFN-Nest, CrossFuse, MUFusion, and EMMA. Among them, RFN-Nest and CrossFuse exhibit a bias toward the visible modality, leading to blurred edges of infrared targets. While the memory unit in MUFusion enhances fusion consistency, it propagates noise, as observed in the human targets in Fig. \ref{TNO}. EMMA relies on unsupervised cross-modal consistency constraints but lacks explicit supervision for edge details, as highlighted in the blue box in Fig. \ref{TNO}. The fourth group consists of task-driven methods, including SemLA, ReCoNet, and GIFNet. These approaches overly rely on high-level semantic or functional features, suppressing visible-light details such as the license plate of the car in Fig. \ref{MSRS}. In contrast to these methods, our approach successfully integrates the thermal radiation information from the infrared modality with the texture details from the visible modality, preserving complex features in highly correlated regions while clearly separating salient targets in low-correlation regions, achieving an optimal balance.

\subsubsection{Quantitative comparison}
We conducted a quantitative analysis of the proposed method using six metrics, as shown in Tab. \ref{Quantitave TNO and MSRS}. Our method exhibits significant performance improvements on nearly all metrics, confirming that it is universally applicable to various scenarios, capable of generating images consistent with human visual perception while retaining more complementary information. However, compared to the latest methods, our approach does not achieve the highest scores in sharpness-related metrics (AG, SF). This is due to the introduced noise in their results, which leads to inflated AG and SF values. In contrast, our method preserves more comprehensive information overall while simultaneously reducing noise interference.

\subsection{Ablation studies}
In this section, we analyse each key component of GrFormer, including: the spatial and channel attention modules, cross-modal fusion strategy, cross-modal attention mechanism, manifold network layer configuration, and a comparison with Euclidean-based methods.

\begin{figure}[t]
    \captionsetup{justification=justified,singlelinecheck=false}
        \centering
        \includegraphics[width=1\linewidth]
        {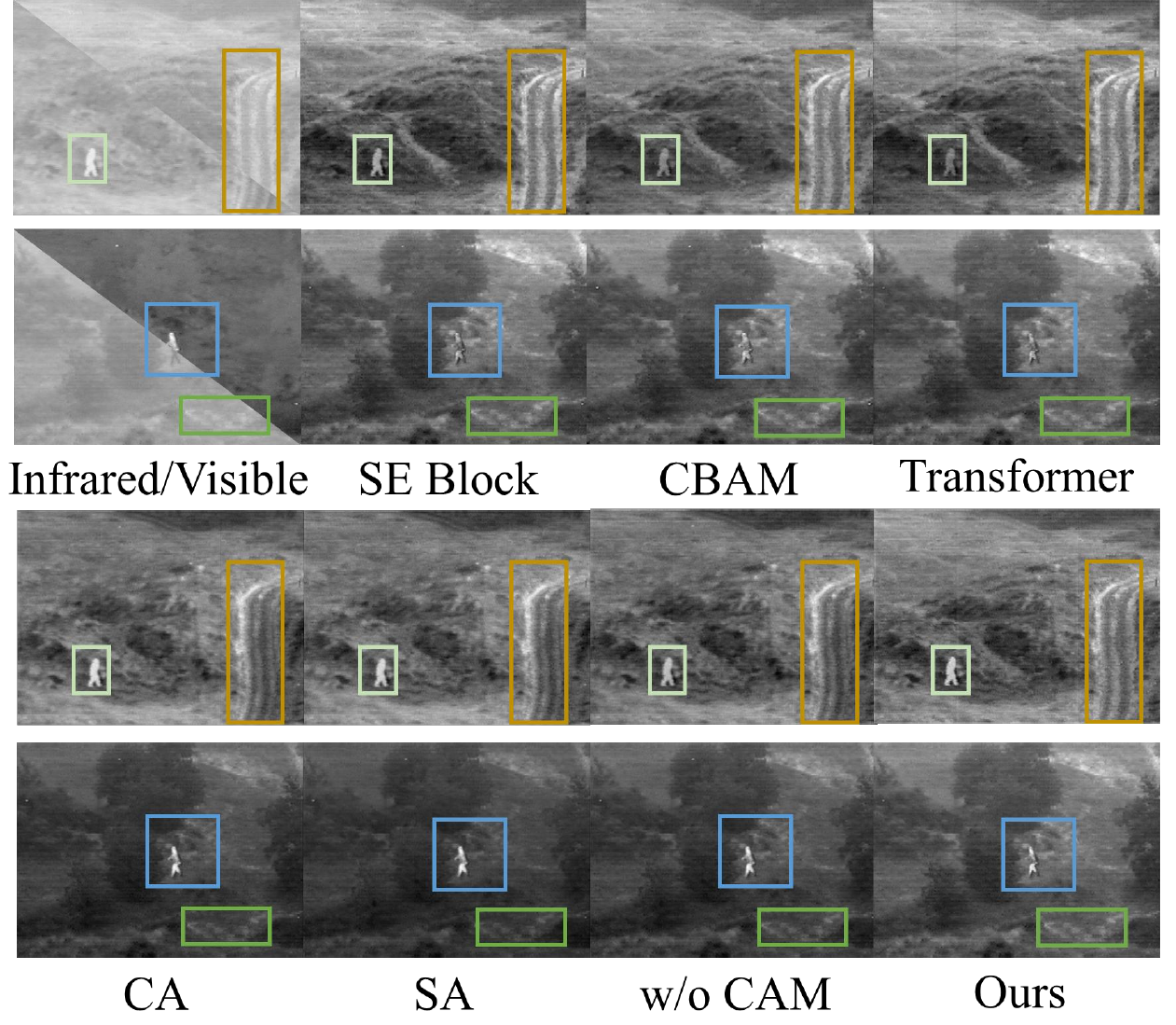}
        \label{fig:visible}

    \caption{ Results of ablation study in different environments. Compared to traditional Euclidean attention mechanisms, our method successfully separates low-frequency semantics from the background. Meanwhile, the hybrid attention manifold network based on channel and spatial dimensions suppresses redundant information. Furthermore, the use of the cross attention mechanism preserves more high-frequency details in RGB images.}
    \label{Ablation1}
\end{figure}

\subsubsection{The influence of ``SA" block and ``CA" block}
Similar to the traditional CBAM~\cite{woo2018cbam}, our network architecture also incorporates attention operations in both the channel and spatial dimensions, which helps the model adaptively emphasize those feature channels and key regions that are more important for the fusion task.

Channel attention focuses on the most important channels in the feature map for the current task and assigns weights to them. As shown in Fig. \ref{Ablation1}, the person in the field might be the part that ``needs to be emphasized", but the model overlooks the texture in the edge areas. In contrast, spatial attention focuses on some important spatial locations but loses the interdependencies between channels, leading to color distortion in the image. Our GrFormer, while balancing the relationship between the two, generates clear fused images.

\begin{table}[ht!]
\centering 
\small
\caption{The average value of the objective metrics obtained using the CA or SA block on the TNO dataset. The best results are highlighted in \textbf{BOLD} fonts.}

\label{CASA}
\begin{tabular}{ p{1.5cm} p{1.2cm} p{1.2cm} p{1.2cm} } 
\toprule
   CA  & \checkmark & & \checkmark\\
   SA  & & \checkmark & \checkmark\\
\midrule
 MI & 3.453 & 3.467 & \textbf{4.023} \\
 SF & 9.484 & 8.758 & \textbf{11.100}\\
 VIF & 0.754 & 0.733 & \textbf{0.873}\\
 AG & 3.711 & 3.485 & \textbf{4.261}\\
 \(Q^{AB/F}\) & 0.486 & 0.472 & \textbf{0.538}\\
 SSIM & 0.491 & \textbf{0.492} & 0.481\\

\bottomrule
\end{tabular}
\end{table}

We first separately trained the spatial and channel attention modules as the backbone of our network. Tab. \ref{CASA} shows that although the outcomes based on CA are slightly higher than those based on SA, the model lacks the ability to localize spatial features, which is not conducive to the fusion of pixel-level complementary information. 
Experiments demonstrate that the training strategy of using both SA and CA can enhance the representational power of pre-fusion features and improve the robustness of training.

\begin{figure}[ht]
    \centering\captionsetup{justification=justified,singlelinecheck=false}
        \centering
        \includegraphics[width=1\linewidth]
        {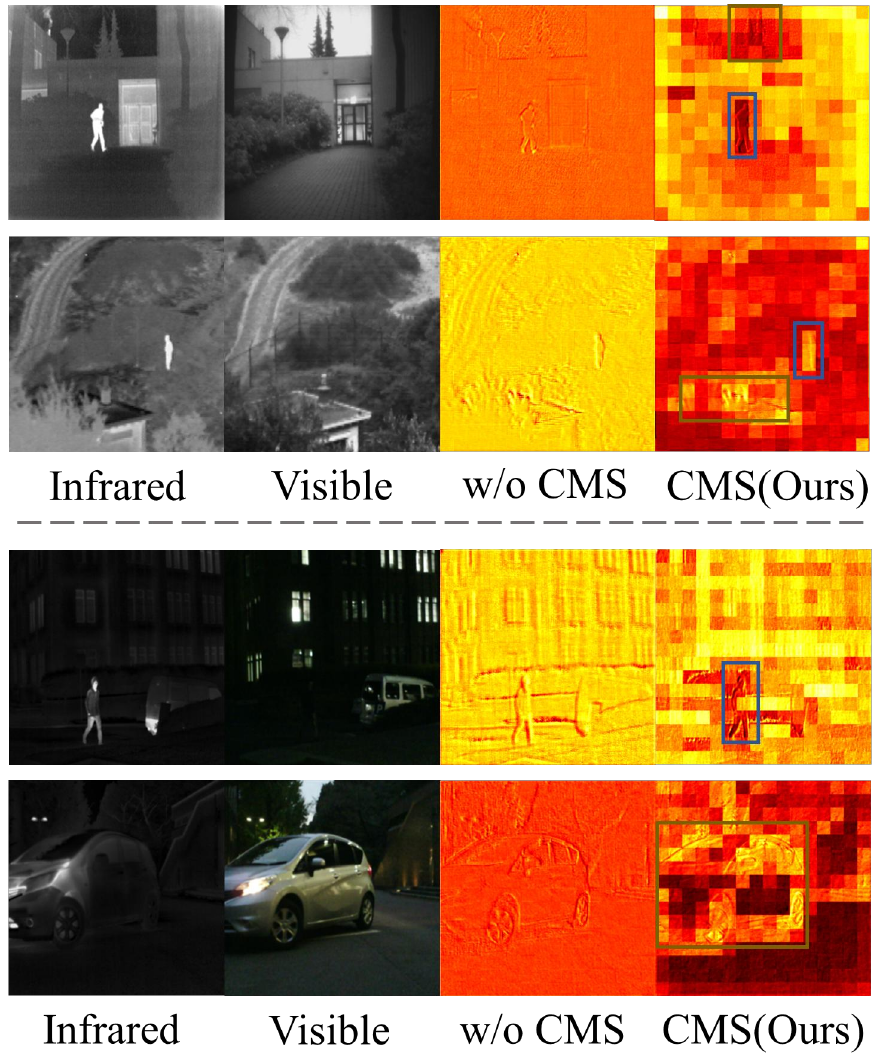}
        \label{fig:visible}
    \caption{ Comparison of intermediate feature visualisation with and without CMS integration. Our method highlights the low-correlation regions between modalities, which are crucial for the fusion task. At the same time, the high-correlation regions are appropriately attenuated, achieving effective suppression of redundant information, thereby enhancing the quality of the fusion results.}
    \label{CMS}
\end{figure}

\subsubsection{The influence of CMS}
Unlike traditional attention guidance methods, we innovatively added a cross-modal mask strategy after the projection matrix, aiming to force the network to learn deep statistical information across modalities. We sequentially incorporated and removed the CMS strategy in our network to evaluate its effectiveness. As shown in Fig. \ref{CMS}, conventional attention operations assign higher weights to background noise or irrelevant information, leading to the insufficient distinction between inter-modal complementary information and redundant information. In contrast, our method highlights salient targets and local textures, preserving the low- correlation detail parts of different modalities. This information may be crucial for distinguishing different objects or scenes.

As shown in Tab. \ref{CMSGSCM}, the fusion results obtained by our method are clearer and more natural, preserving the attention-worthy details from different modalities and enhancing complementary features.

\begin{table}[ht!]
\centering 
\small
\caption{The average value of the objective metrics achieved on the TNO dataset with or without CMS or GSCM. The best results are highlighted in \textbf{BOLD} fonts.}

\label{CMSGSCM}
\begin{tabular}{ p{1.5cm} p{1.2cm} p{1.2cm} p{1.2cm} } 
\toprule
   CMS  & & & \checkmark\\
   GSCM  & & \checkmark & \checkmark\\
\midrule
 MI & 3.153  & 3.200 & \textbf{4.023}\\
 SF & 9.600  & 9.684 & \textbf{11.100}\\
 VIF & 0.745  & 0.731 & \textbf{0.873}\\
 AG & 3.771  & 3.755 & \textbf{4.261}\\
 \(Q^{AB/F}\) & 0.485  & 0.479 & \textbf{0.538}\\
 SSIM & \textbf{0.498}  & 0.495 & 0.481\\

\bottomrule
\end{tabular}
\end{table}

\subsubsection{The influence of CAM}
Our cross-modal attention module incorporates the proposed CMS, which is designed to enhance the complementary information between different modalities. To reveal the rationality of the cross-modal attention network on the manifold, we removed the last two cross-modal manifold networks and trained only the first two self-attention manifold layers (GSSM-SA, GSSM-CA). As illustrated in Fig. \ref{Ablation1}, CAM stands for cross-modal attention module.  Obviously, compared to the results obtained from the fusion network without CAM, GrFormer retains more details, indicating that the texture parts in the visible images are emphasized, while the infrared thermal radiation information is fully preserved, resulting in clearer images. Moreover, thanks to the CMS operation, this complementary information is further amplified, achieving high-quality multi-modal image fusion.

Tab. \ref{CMSGSCM} shows that compared with the network with CAM added, the significance of the target in the fusion result learned by the network without the cross-modal manifold module is significantly reduced. This indicates that in the self-attention mechanism, the model mainly focuses on the information interaction within its own modality, but the mining of the correlation between different modalities is insufficient, and some pixel intensity information in the infrared modality is lost, resulting in an unsatisfactory fusion result.

\begin{figure}[t]
    \centering\captionsetup{justification=justified,singlelinecheck=false}
        \centering
        \includegraphics[width=1\linewidth]
        {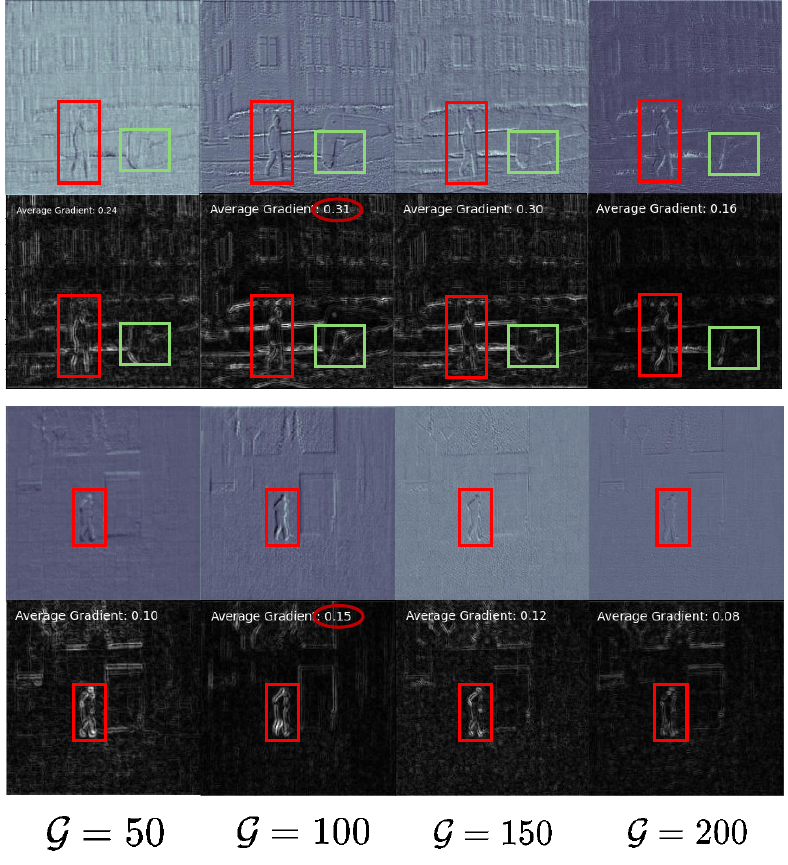}
        \label{fig:visible}
    \caption{ Visualization of semantic feature maps weighted by Grassmann manifolds at different scales. When the subspace coefficient is set to 100, the topological structure of the image is well-preserved while encapsulating rich semantic information. In quantitative evaluations, our method achieves the best performance.}
    \label{Grassmann Attention}
\end{figure}

\begin{table}[ht!]
\centering 
\small
\caption{The average value of the objective metrics achieved with different subspace coefficient \(q\) on the TNO dataset. The best results are highlighted in \textbf{BOLD} fonts.}

\label{q}
\begin{tabular}{ p{1.2cm} p{1cm} p{1cm} p{1cm} p{1cm} } 
\toprule
   \(q\)  & 50 & 100 & 150 & 200  \\
\midrule
 MI & 3.664 & \textbf{4.023} & 3.294 & 3.407\\
 SF & 8.813 & \textbf{11.100} & 9.505 & 9.321\\
 VIF & 0.755 & \textbf{0.873} & 0.759 & 0.756\\
 AG & 3.488 & \textbf{4.261} & 3.697 & 3.679\\
 \(Q^{AB/F}\) & 0.476 & \textbf{0.538} & 0.492 & 0.488\\
SSIM & 0.479 & 0.481 & 0.491 & \textbf{0.492}\\

\bottomrule
\end{tabular}
\end{table}

\begin{figure}[t]
    \captionsetup{justification=justified,singlelinecheck=false}
        \centering
        \includegraphics[width=1\linewidth]
        {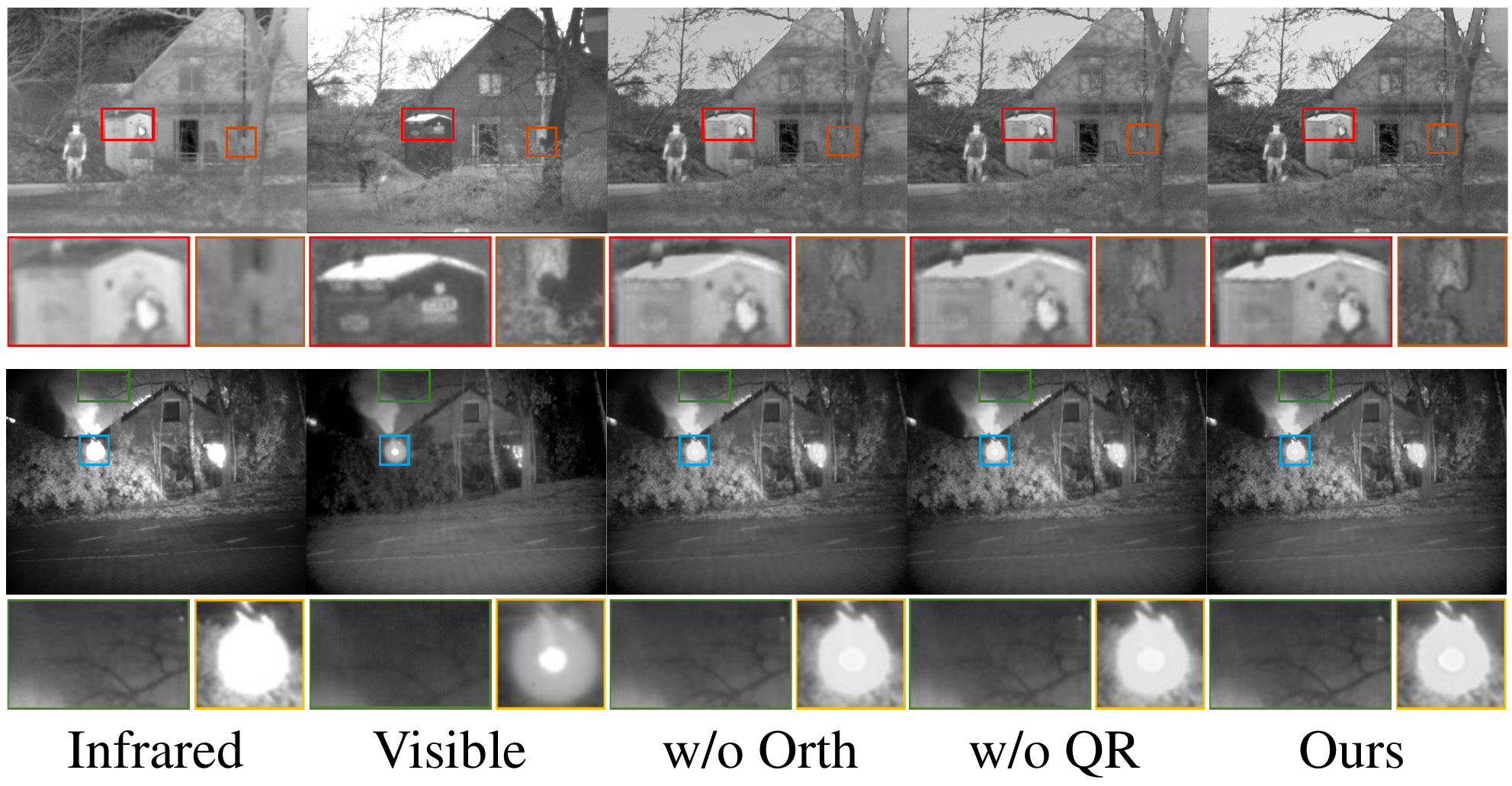}
        \label{fig:visible}

    \caption{ Ablation results under different manifold constraints. Removing the orthogonalization constraint and QR decomposition causes the feature space to deviate from the Grassmann manifold. As shown in the first row, the fused images exhibit color degradation and weakened edge information. The results in the second row fail to clearly distinguish between the center and boundaries of the light source. Our design performs well in both aspects.}
    \label{Ablation3}
\end{figure}

\begin{table}[ht!]
\centering 
\small
\caption{Comparison of average metrics for different manifold constraints on the TNO dataset. The best results are highlighted in \textbf{BOLD} fonts.}

\label{manifold constraints}
\begin{tabular}{ p{1.5cm} p{1.2cm} p{1.2cm} p{1.2cm} } 
\toprule
   Orth & \checkmark & & \checkmark\\
   QR & & \checkmark & \checkmark\\
\midrule
 MI & 3.497  & 3.421 & \textbf{4.023}\\
 SF & 10.270  & 9.789 & \textbf{11.100}\\
 VIF &  0.780 & 0.772 & \textbf{0.873}\\
 AG & 3.991 & 3.821 & \textbf{4.261}\\
 \(Q^{AB/F}\) & 0.502 & 0.499 & \textbf{0.538} \\
 SSIM & \textbf{0.499} & 0.492 & 0.481 \\

\bottomrule
\end{tabular}
\end{table}

\begin{table}[t]
\centering 
\small
\caption{Our module is compared with three Euclidean attention modules. Here, GrAM (Grassmann-based Attention Module) is composed of four attention layers from GrFormer: GSSM-Channel, GSSM-Spatial, GSCM-Channel, and GSCM-Spatial. The best results are highlighted in \textbf{BOLD} fonts.}

\label{Attention Methods}
\begin{tabular}{ p{1.2cm} p{1.2cm} p{1.2cm} p{1.7cm} p{1.2cm} } 
\toprule

     & SE & CBAM & Transformer & GrAM  \\
\midrule

 MI & 3.264 & 3.271 & 3.853 & \textbf{4.023}\\
 SF & 10.958 & 10.418 & 10.938 & \textbf{11.100} \\
 VIF & 0.810 & 0.781 & 0.837 & \textbf{0.873} \\
 AG & 4.254 & 4.135 & \textbf{4.300} & 4.261\\
 \(Q^{AB/F}\) & 0.525 & 0.510 & 0.530 & \textbf{0.538} \\
SSIM & \textbf{0.496} & 0.485 & 0.487 & 0.481\\

\bottomrule
\end{tabular}
\end{table}

\begin{table}[ht!]
\centering 
\small
\caption{Quantitative evaluation of object detection using the MSRS dataset. The best three metrics are highlighted in \textcolor{red}{\textbf{RED}}, \textcolor{brown}{\textbf{BROWN}}, and \textcolor{blue}{\textbf{BLUE}} fonts, respectively.}    
\label{Quantitative detection}
\begin{tabular}{p{2.2cm} p{0.7cm} p{0.7cm} p{0.7cm} p{0.95cm} p{1.3cm}} 
\toprule
Methods & Person & Car & All & mAP@0.5 & mAP@0.95\\
\midrule
FusionGAN~\cite{ma2019fusiongan}  & 0.643 & \textcolor{brown}{\textbf{0.857}} & 0.750  & 0.764 & 0.443 \\
RFN-Nest~\cite{li2021rfn}  & 0.760 & 0.633 & 0.696  & 0.592 & 0.307 \\
GANMcC~\cite{ma2020ganmcc}  & 0.609 & \textcolor{red}{\textbf{0.883}} & 0.746  & 0.784 & 0.424 \\
ReCoNet~\cite{huang2022reconet}  & 0.676 & \textcolor{blue}{\textbf{0.856}} & 0.766  & 0.681 & 0.388 \\
DeFusion~\cite{liang2022fusion} & \textcolor{brown}{\textbf{0.937}} & 0.798 & \textcolor{brown}{\textbf{0.868}} & \textcolor{red}{\textbf{0.805}} & \textcolor{blue}{\textbf{0.453}}\\
MUFusion~\cite{cheng2023mufusion}  & 0.544 & 0.687 & 0.615  & 0.716 & 0.406 \\
SemLA~\cite{xie2023semantics}  & 0.857 & 0.771 & 0.814 & 0.732 & 0.430 \\
LRRNet~\cite{li2023lrrnet}  & 0.577 & 0.842 & 0.709 & 0.721 & 0.404 \\
CrossFuse~\cite{li2024crossfuse}  & \textcolor{blue}{0.926} & 0.767 & \textcolor{blue}{0.846}  & \textcolor{brown}{\textbf{0.803}} & \textcolor{brown}{\textbf{0.473}} \\
VDMUFusion~\cite{shi2024vdmufusion} & 0.750 & 0.514 & 0.632 & 0.760 & 0.439 \\
EMMA~\cite{zhao2024equivariant}  & 0.852 & 0.793 & 0.822 & 0.785 & 0.430 \\
FusionBooster~\cite{cheng2024fusionbooster} & 0.619 & 0.704 & 0.661 & 0.774 & 0.422 \\
GIFNet~\cite{cheng2025cvpr_gifnet} & 0.512 & 0.740 & 0.626 & 0.693 & 0.388 \\
\midrule
\small
GrFormer & \textcolor{red}{\textbf{0.999}} & 0.788 & \textcolor{red}{\textbf{0.893}} & \textcolor{blue}{\textbf{0.790}} & \textcolor{red}{\textbf{0.492}} \\
\bottomrule
\end{tabular}
\end{table}

    


\begin{figure*}[ht!]
    \small
    \centering\captionsetup{justification=justified,singlelinecheck=false}

        \centering
        \includegraphics[width=1\linewidth]
        {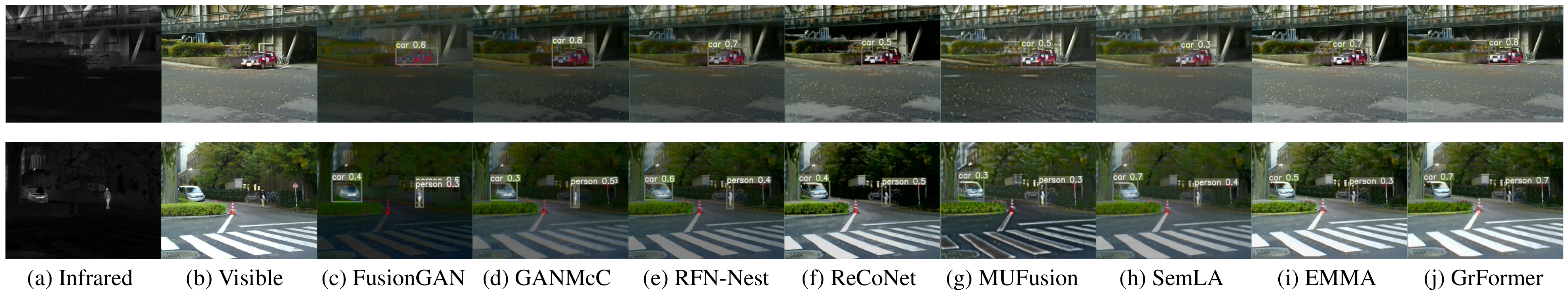}
        \label{fig:visible}

    \caption{ We select the representative MSRS dataset to validate our method. In comparison with other fusion networks, our GrFormer effectively integrates complementary information from infrared and visible images, achieving the highest detection accuracy.}
    \label{detection}
\end{figure*}

\begin{figure*}[ht!]
    \small
    \centering\captionsetup{justification=justified,singlelinecheck=false}

        \centering
        \includegraphics[width=1\linewidth]
        {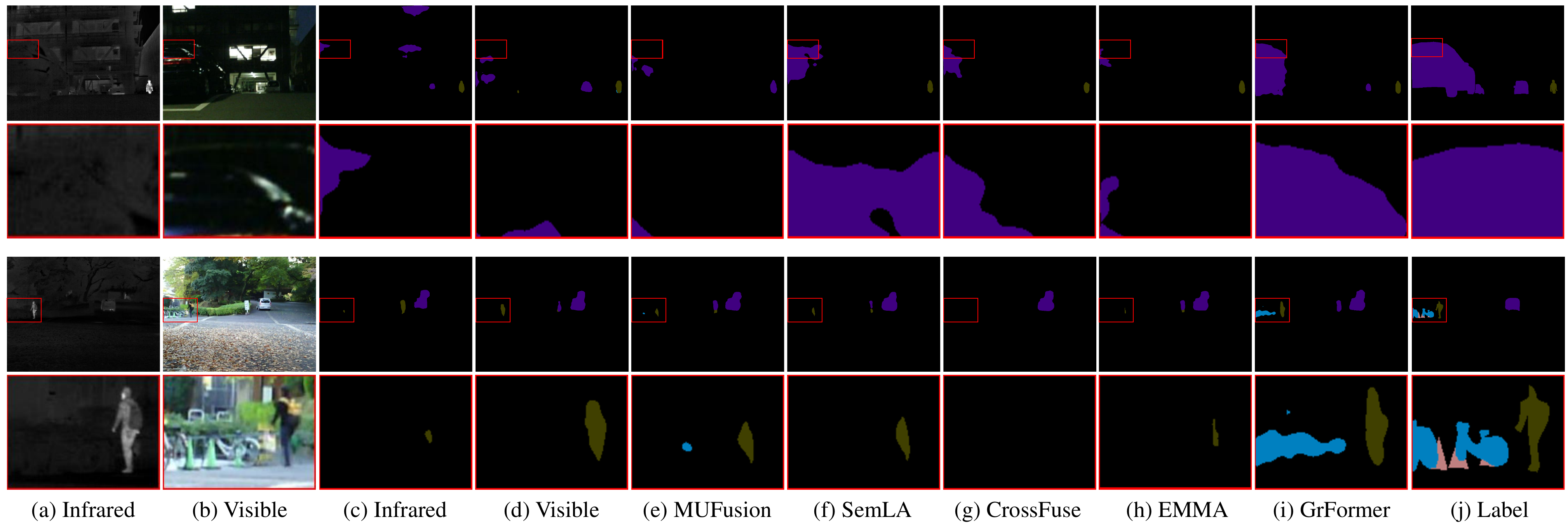}
        \label{fig:visible}

    \caption{ We conducted comparative segmentation experiments on the MSRS dataset. As demonstrated in the examples, GrFormer is capable of effectively segmenting thermally insensitive objects (such as bicycles on the roadside). For objects with high thermal information content (such as cars and people), our method fully leverages these thermal cues, generating more desirable segmentation results.}
    \label{segmentation}
\end{figure*}

\subsubsection{Analysis of the projection coefficient}
During the process of projecting the network into different Grassmann subspaces, we aggregate information across all channel dimensions to obtain a multi-scale low-rank representation. However, the subspace representation based on the spatial dimension incurs a significant computational cost due to large-scale EIG and QR decompositions. Therefore, to better perform representation learning, we conducted ablation studies on subspaces of different scales to identify the optimal experimental setup.

We set subspace coefficients in the FRMap layer to 50, 100, 150, and 200, respectively, and we visualized the results.
As shown in Fig. \ref{Grassmann Attention}, when \(\mathcal{G}\)=50, the feature map contains fewer detailed information and may fail to capture complex data structures. The pedestrian is almost indistinguishable from the background, losing some of the shape features of the car. At \(\mathcal{G}\)=150, the attention network retains some texture details from the visible images, but introduces noise. When \(\mathcal{G}\)=200, the computational efficiency significantly decreases, severe distortion occurs at the edges, and the distinction between pedestrians and the background is greatly reduced. In our method, we set \(\mathcal{G}\)=100, achieving a balance between feature representation capability and computational efficiency. At this point, the image highlights the pedestrian features while preserving the texture details of the background, achieving enhancement in semantic information from both modalities.

As displayed in Tab. \ref{q}, when the subspace coefficients are set too low, their dimensionality becomes insufficient to characterize the high-curvature geometric characteristics of the manifold, leading to aggravated local geometric distortions and significant degradation in image details and topological structures. When the subspace coefficients are set too high, the computational complexity of high-dimensional matrix decomposition grows exponentially, and redundant dimensions introduce spurious curvature noise, resulting in the loss of complementary information. Therefore, we set a moderate coefficient value to achieve the best fused visual effects.

\begin{table}[ht!]
\centering 
\caption{Quantitative evaluation of segmentation using the MSRS dataset. The best three metrics are highlighted in \textcolor{red}{\textbf{RED}}, \textcolor{brown}{\textbf{BROWN}}, and \textcolor{blue}{\textbf{BLUE}} fonts, respectively.}    
\label{Quantitative segmentation}
\begin{tabular}{p{2.4cm} p{0.8cm} p{0.8cm} p{0.8cm} p{0.8cm} p{0.8cm}} 
\toprule
Methods & Unl & Car & Per & Bik & mIOU  \\
\midrule
Visible & 97.89 & \textcolor{blue}{\textbf{74.75}} & 35.31 & 62.59 & 67.63 \\
Infrared & 96.93 & 61.13 & 15.14 & 45.77 & 54.74 \\
FusionGAN~\cite{ma2019fusiongan} & 97.72 & 71.49 & \textcolor{red}{\textbf{42.09}} & 61.82 & 68.28 \\
RFN-Nest~\cite{li2021rfn} & \textcolor{blue}{\textbf{97.91}} & \textcolor{brown}{\textbf{75.35}} & 31.90 & 61.81 & 66.74 \\
GANMcC~\cite{ma2020ganmcc} & \textcolor{blue}{\textbf{97.91}} & 72.92 & 38.11 & 64.22 & \textcolor{blue}{\textbf{68.29}} \\
ReCoNet~\cite{huang2022reconet} & 97.52 & 70.89 & 25.94 & 54.60 & 62.24 \\
DeFusion~\cite{liang2022fusion} & 97.85 & 74.18 & 21.84 & 64.70 & 64.64 \\
MUFusion~\cite{cheng2023mufusion} & 97.78 & 71.53 & 35.97 & 64.76 & 67.51 \\

SemLA~\cite{xie2023semantics} & 97.84 & 73.87 & 32.96 & 59.95 & 66.16 \\
LRRNet~\cite{li2023lrrnet} & 97.86 & 73.57 & \textcolor{brown}{\textbf{40.06}} & 59.21 & 67.68 \\
CrossFuse~\cite{li2024crossfuse} & 97.86 & 73.70 & 33.00 & 62.07 & 66.66 \\
VDMUFusion~\cite{shi2024vdmufusion} & 97.80 & 73.10 & 32.92 & 59.07 & 65.73 \\
EMMA~\cite{zhao2024equivariant} & \textcolor{brown}{\textbf{97.94}} & 73.99 & \textcolor{blue}{\textbf{38.81}} & \textcolor{blue}{\textbf{64.91}} & \textcolor{brown}{\textbf{68.91}} \\
FusionBooster~\cite{cheng2024fusionbooster} & 97.87 & 72.78 & 34.38 & \textcolor{brown}{\textbf{65.30}} & 67.58 \\
GIFNet~\cite{cheng2025cvpr_gifnet} & 97.85 & 73.19 & 35.98 & 61.30 & 67.08 \\
\midrule
GrFormer & \textcolor{red}{\textbf{98.15}} & \textcolor{red}{\textbf{78.42}} & 36.03 & \textcolor{red}{\textbf{66.77}} & \textcolor{red}{\textbf{69.84}} \\
\bottomrule
\end{tabular}
\end{table}

\subsubsection{Analysis of the manifold constraints}
The core of the Grassmann manifold lies in maintaining the stability and directional consistency of feature subspaces through orthogonality. Orthogonalization constraints ensure the orthogonality of the initial mapping matrix, preventing redundant or ill-conditioned structures in the feature space during decomposition. To validate the effectiveness of the Grassmann manifold network, we replaced the initial orthogonalization constraint with a random mapping matrix and eliminated QR decomposition. As shown in the first row of Fig. \ref{Ablation3}, removing the orthogonalization constraint leads to chaotic feature directions, resulting in insufficient subspace decomposition, which in turn causes color degradation and edge blurring in the fused image. The role of QR decomposition is to dynamically correct the feature space during optimization, counteracting feature drift caused by numerical errors. When this mechanism is removed, the feature subspace gradually deviates from the orthogonal structure during training, making it difficult for the model to accurately model lighting distribution and texture details, as demonstrated in the second row.

The average values of the six metrics are shown in Tab. \ref{manifold constraints}. Compared with the methods that remove the two manifold constraints, our approach demonstrates superior performance in most metrics, indicating that Grassmann manifold constraints stabilize feature space structure and enhance fusion quality.

\subsubsection{Comparative analysis of manifold-based versus Euclidean attention modules}

We conducted comparative experiments on the proposed Grassmann Attention Module, selecting several classical Euclidean attention mechanisms to validate the performance of our method. These include the spatial domain-based SE Block~\cite{hu2018squeeze}, the combined channel and spatial domain CBAM~\cite{woo2018cbam}, and the classical Transformer architecture~\cite{dosovitskiy2020vit}. Specifically, we replaced the four-layer Grassmann-based Transformer Module in Fig. \ref{workflow of GrFormer} (a) with the aforementioned Euclidean attention modules.
The results of image fusion are illustrated in Fig. \ref{Ablation1}. Compared to other architectures, our method demonstrates superior performance in visual effects, effectively preserving salient infrared features and scene details. Furthermore, the superior metrics in Tab. \ref{Attention Methods} corroborate this observation.

\subsection{Experiments in object detection}


To evaluate the detection performance of fused images, we trained each fusion method’s output on the MSRS dataset~\cite{tang2022piafusion} using YOLOv7~\cite{wang2023yolov7} as the detection network. The evaluation metrics included accuracy, mean precision at 0.5 (AP50), and mean precision at 0.5:0.95 (AP50:95). For the training setup, we configured the following parameters: a batch size of 16, 50 training epochs, 8 dataloader workers, and 2 detection categories (``Person" and ``Car"), where ``All" denotes their average accuracy. All input images were resized to 640×640, and the Adam optimizer was employed for parameter updates.

As shown in Fig. \ref{detection}, among the methods we compared, FusionGAN and GANMcC exhibited redundant detections, failing to accurately distinguish the targets. ReCoNet, MUFusion, and SemLA methods encountered difficulties in detecting the ``car" category, resulting in lower accuracy. Additionally, the results from RFN-Nest and EMMA did not accurately detect pedestrians on the road. In contrast, our GrFormer maintained high detection accuracy in challenging scenarios while preserving the significant features and texture details of the targets.

In terms of quantitative performance, as shown in Tab. \ref{Quantitative detection}, GrFormer has the best detection performance, especially in the ``Person", ``All" and ``AP50:95" categories, indicating that GrFormer can highlight infrared thermal radiation information and adaptively adjust environmental brightness to improve detection accuracy.

\subsection{Experiments in semantic segmentation}

To further validate the performance of the proposed GrFormer in downstream tasks, we conducted comparative experiments using the segmentation network DeeplabV3+~\cite{chen2018encoder} on the aforementioned 13 fusion methods. Specifically, we performed semantic segmentation on the four basic categories (car, person, bike, and background) provided by the MSRS segmentation dataset. The quantitative results were calculated using the average accuracy and mIoU (Mean Intersection over Union).

We train the segmentation network with SGD for 50 epochs on 480×640 inputs, using an initial learning rate of 7e-3 and a batch size of 8. Moreover, we employ automatic LR scaling based on batch size to maintain training stability and accelerate convergence.

As shown in Tab. \ref{Quantitative segmentation}, our method achieved the best scores across four metrics, demonstrating its advantages in enhancing both the overall target regions and detailed boundaries. This also proves that our Grassmann-based fusion method achieves a balanced optimization of global semantics and local details.

Fig. \ref{segmentation} illustrates the comparison with other competing fusion schemes. Clearly, for heat-insensitive objects such as the bikes in the last two rows, our method effectively preserves the basic shapes and detailed information. Meanwhile, for infrared targets, we also highlight their salient features, as shown in the car examples in the first two rows and the person examples in the last two rows.

In summary, both quantitative and qualitative results demonstrate the strong competitiveness of our Grassmann-based attention network.

\begin{table}[t]
\centering 
\small
\caption{Efficiency comparison between GrFormer and 13 SOTA methods. The best results are highlighted in \textbf{BOLD} fonts.}

\label{efficiency comparison}
\begin{tabular}{ p{3cm} p{2.2cm} p{2.2cm}} 
\toprule

   Methods  & Params(MB) & FLOPs(G) \\
\midrule
FusionGAN~\cite{ma2019fusiongan} & 0.930 & 496.800 \\
RFN-Nest~\cite{li2021rfn} & 2.733 & 89.797 \\
GANMcC~\cite{ma2020ganmcc} & 1.864 & 1002.560 \\
ReCoNet~\cite{huang2022reconet} & \textbf{0.007} & \textbf{1.518} \\
DeFusion~\cite{liang2022fusion} & 7.874 & 15.265 \\
MUFusion~\cite{cheng2023mufusion} & 0.555 & 72.201 \\
SemLA~\cite{xie2023semantics} & 7.278 & 16.613 \\
LRRNet~\cite{li2023lrrnet} & 0.049 & 14.167 \\
CrossFuse~\cite{li2024crossfuse} & 19.212 & 41.516 \\
VDMUFusion~\cite{shi2024vdmufusion} & 31.751 & 321.972 \\
EMMA~\cite{zhao2024equivariant} & 1.516 & 41.538 \\
FusionBooster~\cite{cheng2024fusionbooster} & 0.555 & 56.407 \\
GIFNet~\cite{cheng2025cvpr_gifnet} & 0.613 & 39.814 \\
\midrule
GrFormer & 0.922 & 71.022 \\
\bottomrule
\end{tabular}
\end{table}

\subsection{Efficiency comparison}
Tab. \ref{efficiency comparison} presents a computational efficiency comparison between GrFormer and 13 other methods, evaluated using both parameter count (Params) and floating-point operations (FLOPs). Notably, GAN-based fusion methods typically introduce substantial computational overhead. Methods like CrossFuse, EMMA, and SemLA incorporate Vision Transformer (ViT) architectures, resulting in increased parameter counts. Other approaches such as DeFusion employ complex feature decomposition modules, RFN-Nest adopts a two-stage training strategy, and MUFusion integrates memory units - all contributing additional computational costs.
In contrast, lightweight designs in ReCoNet, LRRNet, FusionBooster, and GIFNet achieve relatively lower parameter counts and computational requirements. Compared to these methods, GrFormer's runtime performance is less competitive due to its transformer architecture and the CPU-dependent eigenvalue decomposition operations in its manifold network, which impact time efficiency. Nevertheless, GrFormer's simple hierarchical structure design enables it to surpass most existing methods in terms of parameter efficiency.

\section{Conclusion}
This paper proposes a Grassmann manifold embedding attention architecture for multimodal fusion. We generalize the manifold attention paradigm to achieve semantic similarity computation in both channel and spatial domains. Furthermore, we extend the self-attention-based manifold learning network to a cross-modal fusion network and innovatively design a mask strategy to enhance the network's focus on complementary features between different modalities. The multi-task experimental results demonstrate that our model not only excels in fusion tasks but also achieves encouraging results in downstream detection tasks.

While the current Grassmann manifold-based attention model performs well in multimodal fusion, its high computational cost may limit practical applications. Future work will focus on developing efficient lightweight versions using techniques like low-rank approximation and dynamic sparsification. We also plan to incorporate additional modalities (e.g., the text modality) through cross-modal alignment methods to improve heterogeneous data understanding.

\section*{Acknowledgement}

This work was supported by the National Natural Science Foundation of China (62020106012, U1836218, 62106089, 62202205), the 111 Project of Ministry of Education of China (B12018), the Engineering and Physical Sciences Research Council (EPSRC) ( EP/V002856/1).


%









{
    \small
    \bibliographystyle{elsarticle-num}
    \bibliography{main}

\begin{thebibliography}{10}
\expandafter\ifx\csname url\endcsname\relax
  \def\url#1{\texttt{#1}}\fi
\expandafter\ifx\csname urlprefix\endcsname\relax\def\urlprefix{URL }\fi
\expandafter\ifx\csname href\endcsname\relax
  \def\href#1#2{#2} \def\path#1{#1}\fi

\bibitem{ma2019infrared}
J.~Ma, Y.~Ma, C.~Li, Infrared and visible image fusion methods and applications: A survey, Information fusion 45 (2019) 153--178.

\bibitem{zhang2021image}
H.~Zhang, H.~Xu, X.~Tian, J.~Jiang, J.~Ma, Image fusion meets deep learning: A survey and perspective, Information Fusion 76 (2021) 323--336.

\bibitem{jiang2021review}
X.~Jiang, J.~Ma, G.~Xiao, Z.~Shao, X.~Guo, A review of multimodal image matching: Methods and applications, Information Fusion 73 (2021) 22--71.

\bibitem{xu2020u2fusion}
H.~Xu, J.~Ma, J.~Jiang, X.~Guo, H.~Ling, U2fusion: A unified unsupervised image fusion network, IEEE Transactions on Pattern Analysis and Machine Intelligence 44~(1) (2020) 502--518.

\bibitem{prakash2021multi}
A.~Prakash, K.~Chitta, A.~Geiger, Multi-modal fusion transformer for end-to-end autonomous driving, in: Proceedings of the IEEE/CVF conference on computer vision and pattern recognition, 2021, pp. 7077--7087.

\bibitem{tang2022matr}
W.~Tang, F.~He, Y.~Liu, Y.~Duan, Matr: Multimodal medical image fusion via multiscale adaptive transformer, IEEE Transactions on Image Processing 31 (2022) 5134--5149.

\bibitem{zhou2023gan}
T.~Zhou, Q.~Li, H.~Lu, Q.~Cheng, X.~Zhang, Gan review: Models and medical image fusion applications, Information Fusion 91 (2023) 134--148.

\bibitem{tang2023exploring}
Z.~Tang, T.~Xu, H.~Li, X.-J. Wu, X.~Zhu, J.~Kittler, Exploring fusion strategies for accurate rgbt visual object tracking, Information Fusion 99 (2023) 101881.

\bibitem{zhu2022visual}
X.-F. Zhu, T.~Xu, X.-J. Wu, Visual object tracking on multi-modal rgb-d videos: a review, arXiv preprint arXiv:2201.09207 (2022).

\bibitem{dosovitskiy2020vit}
A.~Dosovitskiy, L.~Beyer, A.~Kolesnikov, D.~Weissenborn, X.~Zhai, T.~Unterthiner, M.~Dehghani, M.~Minderer, G.~Heigold, S.~Gelly, J.~Uszkoreit, N.~Houlsby, An image is worth 16x16 words: Transformers for image recognition at scale, ICLR (2021).

\bibitem{hu2018squeeze}
J.~Hu, L.~Shen, G.~Sun, Squeeze-and-excitation networks, in: Proceedings of the IEEE conference on computer vision and pattern recognition, 2018, pp. 7132--7141.

\bibitem{vaswani2017attention}
A.~Vaswani, Attention is all you need, Advances in Neural Information Processing Systems (2017).

\bibitem{wang2018non}
X.~Wang, R.~Girshick, A.~Gupta, K.~He, Non-local neural networks, in: Proceedings of the IEEE conference on computer vision and pattern recognition, 2018, pp. 7794--7803.

\bibitem{woo2018cbam}
S.~Woo, J.~Park, J.-Y. Lee, I.~S. Kweon, Cbam: Convolutional block attention module, in: Proceedings of the European conference on computer vision (ECCV), 2018, pp. 3--19.

\bibitem{zhang2019self}
H.~Zhang, I.~Goodfellow, D.~Metaxas, A.~Odena, Self-attention generative adversarial networks, in: International conference on machine learning, PMLR, 2019, pp. 7354--7363.

\bibitem{li2020nestfuse}
H.~Li, X.-J. Wu, T.~Durrani, Nestfuse: An infrared and visible image fusion architecture based on nest connection and spatial/channel attention models, IEEE Transactions on Instrumentation and Measurement 69~(12) (2020) 9645--9656.

\bibitem{rao2023tgfuse}
D.~Rao, T.~Xu, X.-J. Wu, Tgfuse: An infrared and visible image fusion approach based on transformer and generative adversarial network, IEEE Transactions on Image Processing (2023).

\bibitem{xiao2020global}
B.~Xiao, B.~Xu, X.~Bi, W.~Li, Global-feature encoding u-net (geu-net) for multi-focus image fusion, IEEE Transactions on Image Processing 30 (2020) 163--175.

\bibitem{jia2023multiscale}
S.~Jia, Z.~Min, X.~Fu, Multiscale spatial--spectral transformer network for hyperspectral and multispectral image fusion, Information Fusion 96 (2023) 117--129.

\bibitem{li2024crossfuse}
H.~Li, X.-J. Wu, Crossfuse: A novel cross attention mechanism based infrared and visible image fusion approach, Information Fusion 103 (2024) 102147.

\bibitem{kang2024spdfusion}
H.~Kang, H.~Li, T.~Xu, R.~Wang, X.-J. Wu, J.~Kittler, Spdfusion: An infrared and visible image fusion network based on a non-euclidean representation of riemannian manifolds, arXiv preprint arXiv:2411.10679 (2024).

\bibitem{li2018densefuse}
H.~Li, X.-J. Wu, Densefuse: A fusion approach to infrared and visible images, IEEE Transactions on Image Processing 28~(5) (2018) 2614--2623.

\bibitem{zhang2020ifcnn}
Y.~Zhang, Y.~Liu, P.~Sun, H.~Yan, X.~Zhao, L.~Zhang, Ifcnn: A general image fusion framework based on convolutional neural network, Information Fusion 54 (2020) 99--118.

\bibitem{li2021rfn}
H.~Li, X.-J. Wu, J.~Kittler, Rfn-nest: An end-to-end residual fusion network for infrared and visible images, Information Fusion 73 (2021) 72--86.

\bibitem{li2024deep}
H.~Li, J.~Liu, Y.~Zhang, Y.~Liu, A deep learning framework for infrared and visible image fusion without strict registration, International Journal of Computer Vision 132~(5) (2024) 1625--1644.

\bibitem{li2025mulfs}
H.~Li, Z.~Yang, Y.~Zhang, W.~Jia, Z.~Yu, Y.~Liu, Mulfs-cap: Multimodal fusion-supervised cross-modality alignment perception for unregistered infrared-visible image fusion, IEEE Transactions on Pattern Analysis and Machine Intelligence (2025).

\bibitem{tang2022superfusion}
L.~Tang, Y.~Deng, Y.~Ma, J.~Huang, J.~Ma, Superfusion: A versatile image registration and fusion network with semantic awareness, IEEE/CAA Journal of Automatica Sinica 9~(12) (2022) 2121--2137.

\bibitem{ma2019fusiongan}
J.~Ma, W.~Yu, P.~Liang, C.~Li, J.~Jiang, Fusiongan: A generative adversarial network for infrared and visible image fusion, Information fusion 48 (2019) 11--26.

\bibitem{ma2020ddcgan}
J.~Ma, H.~Xu, J.~Jiang, X.~Mei, X.-P. Zhang, Ddcgan: A dual-discriminator conditional generative adversarial network for multi-resolution image fusion, IEEE Transactions on Image Processing 29 (2020) 4980--4995.

\bibitem{zhao2023metafusion}
W.~Zhao, S.~Xie, F.~Zhao, Y.~He, H.~Lu, Metafusion: Infrared and visible image fusion via meta-feature embedding from object detection, in: Proceedings of the IEEE/CVF Conference on Computer Vision and Pattern Recognition, 2023, pp. 13955--13965.

\bibitem{tang2023dual}
Z.~Tang, G.~Xiao, J.~Guo, S.~Wang, J.~Ma, Dual-attention-based feature aggregation network for infrared and visible image fusion, IEEE Transactions on Instrumentation and Measurement 72 (2023) 1--13.

\bibitem{ma2022swinfusion}
J.~Ma, L.~Tang, F.~Fan, J.~Huang, X.~Mei, Y.~Ma, Swinfusion: Cross-domain long-range learning for general image fusion via swin transformer, IEEE/CAA Journal of Automatica Sinica 9~(7) (2022) 1200--1217.

\bibitem{zhao2023cddfuse}
Z.~Zhao, H.~Bai, J.~Zhang, Y.~Zhang, S.~Xu, Z.~Lin, R.~Timofte, L.~Van~Gool, Cddfuse: Correlation-driven dual-branch feature decomposition for multi-modality image fusion, in: Proceedings of the IEEE/CVF conference on computer vision and pattern recognition, 2023, pp. 5906--5916.

\bibitem{qu2022transmef}
L.~Qu, S.~Liu, M.~Wang, Z.~Song, Transmef: A transformer-based multi-exposure image fusion framework using self-supervised multi-task learning, in: Proceedings of the AAAI conference on artificial intelligence, Vol.~36, 2022, pp. 2126--2134.

\bibitem{ma2021stdfusionnet}
J.~Ma, L.~Tang, M.~Xu, H.~Zhang, G.~Xiao, Stdfusionnet: An infrared and visible image fusion network based on salient target detection, IEEE Transactions on Instrumentation and Measurement 70 (2021) 1--13.

\bibitem{SMRNet}
G.~Xiao, X.~Liu, Z.~Lin, R.~Ming, Smr-net: Semantic-guided mutually reinforcing network for cross-modal image fusion and salient object detection, in: Thirty-Eighth AAAI Conference on Artificial Intelligence (AAAI), 2025.

\bibitem{ming2025ssdfusion}
R.~Ming, Y.~Xiao, X.~Liu, G.~Zheng, G.~Xiao, Ssdfusion: A scene-semantic decomposition approach for visible and infrared image fusion, Pattern Recognition 163 (2025) 111457.

\bibitem{xiao2024fafusion}
G.~Xiao, Z.~Tang, H.~Guo, J.~Yu, H.~T. Shen, Fafusion: Learning for infrared and visible image fusion via frequency awareness, IEEE Transactions on Instrumentation and Measurement 73 (2024) 1--11.

\bibitem{li2020mdlatlrr}
H.~Li, X.-J. Wu, J.~Kittler, Mdlatlrr: A novel decomposition method for infrared and visible image fusion, IEEE Transactions on Image Processing 29 (2020) 4733--4746.

\bibitem{li2023lrrnet}
H.~Li, T.~Xu, X.-J. Wu, J.~Lu, J.~Kittler, Lrrnet: A novel representation learning guided fusion network for infrared and visible images, IEEE transactions on pattern analysis and machine intelligence 45~(9) (2023) 11040--11052.

\bibitem{zhang2021exploring}
Q.~Zhang, F.~Wang, Y.~Luo, J.~Han, Exploring a unified low rank representation for multi-focus image fusion, Pattern Recognition 113 (2021) 107752.

\bibitem{liu2012robust}
G.~Liu, Z.~Lin, S.~Yan, J.~Sun, Y.~Yu, Y.~Ma, Robust recovery of subspace structures by low-rank representation, IEEE transactions on pattern analysis and machine intelligence 35~(1) (2012) 171--184.

\bibitem{wright2008robust}
J.~Wright, A.~Y. Yang, A.~Ganesh, S.~S. Sastry, Y.~Ma, Robust face recognition via sparse representation, IEEE transactions on pattern analysis and machine intelligence 31~(2) (2008) 210--227.

\bibitem{huang2018building}
Z.~Huang, J.~Wu, L.~Van~Gool, Building deep networks on grassmann manifolds, in: Proceedings of the AAAI Conference on Artificial Intelligence, Vol.~32, 2018.

\bibitem{sharma2020image}
K.~Sharma, R.~Rameshan, Image set classification using a distance-based kernel over affine grassmann manifold, IEEE transactions on neural networks and learning systems 32~(3) (2020) 1082--1095.

\bibitem{wang2023get}
H.~Wang, Z.~Li, W.~Zhang, Get the best of both worlds: Improving accuracy and transferability by grassmann class representation, in: Proceedings of the IEEE/CVF International Conference on Computer Vision, 2023, pp. 22478--22487.

\bibitem{wang2020graph}
R.~Wang, X.-J. Wu, J.~Kittler, Graph embedding multi-kernel metric learning for image set classification with grassmannian manifold-valued features, IEEE Transactions on Multimedia 23 (2020) 228--242.

\bibitem{harandi2014manifold}
M.~T. Harandi, M.~Salzmann, R.~Hartley, From manifold to manifold: Geometry-aware dimensionality reduction for spd matrices, in: Computer Vision--ECCV 2014: 13th European Conference, Zurich, Switzerland, September 6-12, 2014, Proceedings, Part II 13, Springer, 2014, pp. 17--32.

\bibitem{harandi2017dimensionality}
M.~Harandi, M.~Salzmann, R.~Hartley, Dimensionality reduction on spd manifolds: The emergence of geometry-aware methods, IEEE transactions on pattern analysis and machine intelligence 40~(1) (2017) 48--62.

\bibitem{nguyen2022deep}
N.~D. Nguyen, J.~Huang, D.~Wang, A deep manifold-regularized learning model for improving phenotype prediction from multi-modal data, Nature computational science 2~(1) (2022) 38--46.

\bibitem{li2016multi}
J.~Li, Y.~Wu, J.~Zhao, K.~Lu, Multi-manifold sparse graph embedding for multi-modal image classification, Neurocomputing 173 (2016) 501--510.

\bibitem{edelman1998geometry}
A.~Edelman, T.~A. Arias, S.~T. Smith, The geometry of algorithms with orthogonality constraints, SIAM journal on Matrix Analysis and Applications 20~(2) (1998) 303--353.

\bibitem{ionescu2015training}
C.~Ionescu, O.~Vantzos, C.~Sminchisescu, Training deep networks with structured layers by matrix backpropagation, arXiv preprint arXiv:1509.07838 (2015).

\bibitem{TNO}
T.~Alexander, Tno image fusion dataset (2014).

\bibitem{tang2022piafusion}
L.~Tang, J.~Yuan, H.~Zhang, X.~Jiang, J.~Ma, Piafusion: A progressive infrared and visible image fusion network based on illumination aware, Information Fusion 83 (2022) 79--92.

\bibitem{ma2020ganmcc}
J.~Ma, H.~Zhang, Z.~Shao, P.~Liang, H.~Xu, Ganmcc: A generative adversarial network with multiclassification constraints for infrared and visible image fusion, IEEE Transactions on Instrumentation and Measurement 70 (2020) 1--14.

\bibitem{huang2022reconet}
Z.~Huang, J.~Liu, X.~Fan, R.~Liu, W.~Zhong, Z.~Luo, Reconet: Recurrent correction network for fast and efficient multi-modality image fusion, in: European conference on computer Vision, Springer, 2022, pp. 539--555.

\bibitem{liang2022fusion}
P.~Liang, J.~Jiang, X.~Liu, J.~Ma, Fusion from decomposition: A self-supervised decomposition approach for image fusion, in: European Conference on Computer Vision, Springer, 2022, pp. 719--735.

\bibitem{cheng2023mufusion}
C.~Cheng, T.~Xu, X.-J. Wu, Mufusion: A general unsupervised image fusion network based on memory unit, Information Fusion 92 (2023) 80--92.

\bibitem{xie2023semantics}
H.~Xie, Y.~Zhang, J.~Qiu, X.~Zhai, X.~Liu, Y.~Yang, S.~Zhao, Y.~Luo, J.~Zhong, Semantics lead all: Towards unified image registration and fusion from a semantic perspective, Information Fusion 98 (2023) 101835.

\bibitem{shi2024vdmufusion}
Y.~Shi, Y.~Liu, J.~Cheng, Z.~J. Wang, X.~Chen, Vdmufusion: A versatile diffusion model-based unsupervised framework for image fusion, IEEE Transactions on Image Processing (2024).

\bibitem{zhao2024equivariant}
Z.~Zhao, H.~Bai, J.~Zhang, Y.~Zhang, K.~Zhang, S.~Xu, D.~Chen, R.~Timofte, L.~Van~Gool, Equivariant multi-modality image fusion, in: Proceedings of the IEEE/CVF Conference on Computer Vision and Pattern Recognition, 2024, pp. 25912--25921.

\bibitem{cheng2024fusionbooster}
C.~Cheng, T.~Xu, X.-J. Wu, H.~Li, X.~Li, J.~Kittler, Fusionbooster: A unified image fusion boosting paradigm, International Journal of Computer Vision (2025).

\bibitem{cheng2025cvpr_gifnet}
C.~Cheng, T.~Xu, Z.~Feng, X.~Wu, H.~Li, Z.~Zhang, S.~Atito, M.~Awais, J.~Kittler, et~al., One model for all: Low-level task interaction is a key to task-agnostic image fusion, arXiv preprint arXiv:2502.19854 (2025).

\bibitem{wang2004image}
Z.~Wang, A.~C. Bovik, H.~R. Sheikh, E.~P. Simoncelli, Image quality assessment: from error visibility to structural similarity, IEEE transactions on image processing 13~(4) (2004) 600--612.

\bibitem{wang2023yolov7}
C.-Y. Wang, A.~Bochkovskiy, H.-Y.~M. Liao, Yolov7: Trainable bag-of-freebies sets new state-of-the-art for real-time object detectors, in: Proceedings of the IEEE/CVF conference on computer vision and pattern recognition, 2023, pp. 7464--7475.

\bibitem{chen2018encoder}
L.-C. Chen, Y.~Zhu, G.~Papandreou, F.~Schroff, H.~Adam, Encoder-decoder with atrous separable convolution for semantic image segmentation, in: Proceedings of the European conference on computer vision (ECCV), 2018, pp. 801--818.

\end{thebibliography}
}
\end{document}